\documentclass[sn-mathphys,Numbered,onecolumn]{sn-jnl}

\usepackage{lmodern} 

\usepackage{graphicx}
\usepackage[labelfont=bf]{caption}
\usepackage{subcaption}
\usepackage{natbib}
\usepackage{multirow}
\usepackage{amsmath,amssymb,amsfonts}
\usepackage{amsthm}

\usepackage{xcolor}
\usepackage{textcomp}
\usepackage{manyfoot}
\usepackage{booktabs}
\usepackage{comment}

\usepackage{algorithm}
\usepackage{algorithmicx}
\usepackage{algpseudocode}
\usepackage{threeparttable}
\usepackage{multicol}
\usepackage{bm,csquotes}
\usepackage{adjustbox} 
\usepackage{placeins,float}

\usepackage{enumerate}
\counterwithin{figure}{section}
\graphicspath{{figs1/}}
\usepackage{tikz}
\usetikzlibrary{positioning}

\usepackage{geometry}
\usepackage{glossaries}
\makeglossaries
\newacronym{GWE}{GWE}{Greedy Weighting Ensemble}

\begin{document}

\title[Financial Loan Default Prediction]{An Optimised Greedy-Weighted Ensemble Framework for Financial Loan Default Prediction}



\author[1,2]{\fnm{Ezekiel Nii Noye} \sur{Nortey}}

\author[2]{\fnm{Jones} \sur{Asante-Koranteng}}

\author*[3,4]{\fnm{Marcellin} \sur{Atemkeng}}\email{\href{m.atemkeng@gmail.com}{m.atemkeng@ru.ac.za}}

\author[5]{\fnm{Theophilus} \sur{Ansah-Narh}}

\author[2]{\fnm{David} \sur{Mensah}} 

\author[6]{\fnm{Rebecca} \sur{Davis}}
\author[7]{\fnm{Ravenhill Adjetey} \sur{Laryea}}
\affil[1]{\orgdiv{Department of Statistics and Actuarial Science}, \orgname{University of Ghana},  \postcode{P. O. Box LG 115}, \orgaddress{\street{Legon}, \city{Accra}, \country{Ghana}}}

\affil[2]{\orgdiv{Department of Statistics, Tali Graduate school}, \orgname{Dominion University College},  \postcode{P. O. Box LG 80}, \orgaddress{\street{Legon}, \city{Accra},\country{Ghana}}}

\affil*[3]{\orgdiv{Department of Mathematics}, \orgname{Rhodes University},  \postcode{P.O. Box 94}, \orgaddress{\street{Makhanda, 6140}, \city{Grahamstown}, \country{South Africa}}}

\affil*[4]{\orgdiv{National Institute for Theoretical and Computational Sciences (NITheCS)},  \orgaddress{\street{Stellenbosch 7600}, \country{South Africa}}}

\affil*[5]{\orgdiv{Ghana Space Science and Technology Institute}, \orgname{Ghana Atomic Energy Commission},  \postcode{P. O. Box LG 80}, \orgaddress{\street{Legon}, \city{Accra},\country{ Ghana}}}


\affil[6]{\orgdiv{Department of Mathematics and Actuarial Science}, \orgname{Pentecost University},  \postcode{P.O. Box KN 1739}, \state{Kaneshie, Accra}, \country{Ghana}}

\affil[7]{\orgdiv{Department of Economics and Actuarial Science}, \orgname{University of Professional Studies},  \postcode{P. O. Box LG 149}, \orgaddress{\street{Legon}, \city{Accra}, \country{Ghana}}}


\abstract{
\textcolor{black}{
Accurate prediction of loan defaults is a central challenge in credit risk management, particularly in modern financial datasets characterised by nonlinear relationships, class imbalance, and evolving borrower behaviour. Traditional statistical models and static ensemble methods often struggle to maintain reliable performance under such conditions because they rely on fixed model structures or heuristic weighting strategies.
This study proposes an Optimised Greedy-Weighted Ensemble framework for loan default prediction that dynamically allocates model weights based on empirical predictive performance. The framework integrates multiple machine learning classifiers, with their hyperparameters first optimised using Particle Swarm Optimisation. Model predictions are then combined via a regularised greedy weighting mechanism. At the same time, a neural-network-based meta-learner is employed within a stacked-ensemble architecture to capture higher-order relationships among model outputs.
Experiments conducted on the Lending Club dataset demonstrate that the proposed framework substantially improves predictive performance compared with individual classifiers. The BlendNet ensemble achieved the strongest results with an Area Under the Curve of $0.80$, a macro-average F1-score of $0.73$, and a default recall of $0.81$. Calibration analysis further shows that tree-based ensembles such as Extra Trees and Gradient Boosting provide the most reliable probability estimates (Brier score $0.18$), while the stacked ensemble offers superior ranking capability. Feature analysis using Recursive Feature Elimination identifies revolving utilisation, annual income, and debt-to-income ratio as the most influential predictors of loan default.
These findings demonstrate that performance-driven ensemble weighting can improve both predictive accuracy and interpretability in credit risk modelling. The proposed framework provides a scalable data-driven approach to support institutional credit assessment, risk monitoring, and financial decision-making.
}
}


\maketitle
\section{Introduction}  \label{sec:intro}

In the modern financial landscape, loans play a central role in supporting households, businesses, and broader economic growth \cite{miklaszewska2015role}. 
They facilitate home ownership, capital investment, and entrepreneurial activity, yet simultaneously expose lenders and borrowers to significant risks, particularly the risk of default. Loan defaults occur when borrowers fail to meet repayment obligations, leading to substantial financial losses for lending institutions, including uncollected principal, interest, and fees \citep{ntiamoah2014loan,trautmann2013strategic,white1987personal}. Such losses reduce profitability, weaken capital reserves, and can propagate systemic risks, as evidenced during the 2008 global financial crisis \cite{helleiner2011understanding,friedman2009global}. For borrowers, defaults often result in diminished creditworthiness, legal repercussions, and potential loss of collateral, imposing long-term financial and social hardship \cite{domotor2023peer,peng2023legal,saha2023housing,kurniawati2019legal}. 
Consequently, accurate prediction of loan defaults has become a critical priority for financial institutions seeking to strengthen risk management and improve lending decisions.

Traditionally, credit risk modelling has relied on statistical approaches such as Logistic Regression (LR), valued for their interpretability and regulatory acceptance. However, these models often assume relatively stable data-generating processes and linear relationships between predictors and outcomes. In practice, financial data frequently exhibit complex nonlinear dependencies among borrower characteristics, credit history variables, and macroeconomic factors. As a result, conventional statistical models may struggle to capture the evolving patterns present in modern credit datasets \cite{ntiamoah2014loan,trautmann2013strategic}. These limitations have motivated the increasing adoption of machine learning methods capable of modelling nonlinear relationships and high-dimensional interactions.

Despite their flexibility, machine learning models applied to financial risk prediction face several challenges arising from the dynamic nature of economic systems. Empirical studies show that predictive relationships in financial datasets are inherently non-stationary and may evolve over time due to phenomena such as concept drift, data drift, and market regime shifts \cite{Lu2018,Webb2016,Gama2014}. Concept drift occurs when the relationship between predictor variables (e.g., debt-to-income ratio or credit history) and loan default changes over time. For example, a model trained on data prior to the 2008 financial crisis may learn relationships that differ significantly from those observed during subsequent economic recovery or pandemic-related disruptions \cite{Dietterich2000}. Data drift, in contrast, arises when the distribution of input variables changes due to shifts in borrower demographics, lending policies, or macroeconomic conditions. Additionally, financial markets often transition between different regimes characterised by periods of stability, volatility, expansion, or recession, which can further influence predictive model behaviour \cite{Bucci2022}. 

Under such conditions, the predictive performance of individual models may vary substantially over time. A complex learner such as eXtreme Gradient Boosting (XGBoost) may capture intricate nonlinear patterns during stable periods but become sensitive to noise during volatile market conditions, whereas simpler models such as LR may offer greater robustness but reduced flexibility. Consequently, the most effective predictive model for loan default prediction is rarely static; instead, it may vary depending on prevailing data characteristics and economic conditions \cite{Lu2018,Webb2016}.

Ensemble learning methods have emerged as an effective strategy for improving predictive performance by combining multiple models. Approaches such as Bagging, Boosting, and Stacking aggregate predictions from diverse learners in order to reduce variance, improve robustness, and capture complementary patterns in the data \cite{dietterich2000ensemble}. However, many conventional ensemble methods rely on fixed or heuristic weighting strategies that implicitly assume stable model performance. In dynamic financial environments, such assumptions may be unrealistic. Static weighting schemes may overemphasise models whose predictive power has deteriorated due to recent drift, while underutilising models better suited to current data conditions. Furthermore, stacking approaches often determine model contributions implicitly through meta-learners, which can reduce transparency and limit adaptability to evolving data regimes \cite{Ditzler2015,Gama2014}. 

These limitations highlight a methodological gap in existing ensemble frameworks: most approaches do not continuously evaluate model performance or dynamically adjust model contributions in response to changing data environments. Addressing this limitation is particularly important in credit risk modelling, where borrower behaviour, economic conditions, and lending practices evolve.
To address these challenges, we propose a Greedy Weighting Ensemble (GWE) framework for loan default prediction. The proposed framework maintains a diverse pool of base models and dynamically allocates model weights based on empirical predictive performance. Model performance is evaluated over recent data using metrics such as AUC, balanced accuracy, or cost-sensitive loss functions. A greedy optimisation procedure then assigns higher weights to models demonstrating stronger predictive ability while down-weighting weaker learners. To ensure interpretability and numerical stability, the weights are derived through a softmax transformation of performance scores, resulting in a convex combination of base learners that adapts over time.

Although the individual components employed in this framework, including ensemble learning, hyperparameter optimisation, and neural meta-learning, have been studied in prior literature, the novelty of the present work lies in their structured integration within a performance-driven greedy weighting strategy designed to address practical limitations in credit risk modelling. Specifically, the proposed framework emphasises adaptive model weighting, improved robustness under heterogeneous financial data, and enhanced interpretability of ensemble contributions. This design positions the method as a practical extension of existing ensemble techniques rather than a replacement for established approaches.

In many existing credit risk studies, ensemble methods determine model contributions either through fixed weighting schemes or implicitly through a meta-learner within a stacked architecture. In contrast, the proposed GWE introduces an explicit optimisation layer that estimates convex ensemble weights through a regularised greedy search operating directly on probabilistic model predictions. This mechanism differs from many performance-weighted ensembles that assign weights heuristically or through static optimisation procedures, as the weighting process is formulated as an explicit optimisation problem with regularisation and convexity constraints that produce interpretable model contributions.

Another distinguishing aspect of the framework is the separation between hyperparameter optimisation and ensemble weighting. Hyperparameter optimisation of the base learners is performed prior to ensemble construction using Particle Swarm Optimisation (PSO), ensuring that ensemble weights reflect the predictive capability of optimised learners rather than the behaviour of untuned models. Once the base models are calibrated, the greedy weighting mechanism determines their relative contributions, after which the BlendNet meta-learner operates on both weighted and unweighted ensemble outputs. This hybrid meta-feature representation enables the neural network to capture residual dependencies among model predictions and refine the final probability estimates. 

Consequently, the overall architecture decomposes the modelling pipeline into three distinct stages: PSO-based model calibration, greedy convex ensemble weighting, and neural meta-learning. This separation of model optimisation, ensemble aggregation, and meta-level correction differentiates the framework from conventional stacked ensembles in which model weighting is determined solely within the meta-learning stage.
The key contributions of this work are summarised as follows:

\begin{enumerate}[i.]

\item A dynamic ensemble weighting framework is introduced to adaptively allocate model contributions based on predictive performance. This strategy enables the ensemble to emphasise stronger base learners while reducing the influence of weaker models, improving the balance between detecting default and non-default loans compared with conventional fixed-weight ensembles.

\item Informative borrower attributes are identified using Recursive Feature Elimination (RFE) with tree-based estimators, enabling dimensionality reduction while preserving the most relevant predictors. This step improves model interpretability, reduces computational complexity, and enhances generalisability across heterogeneous borrower profiles.

\item Class imbalance in the Lending Club dataset is addressed through the integration of the Synthetic Minority Over-sampling Technique (SMOTE) and cost-sensitive learning strategies. These mechanisms improve sensitivity to default events while mitigating bias toward the majority non-default class.

\item Particle Swarm Optimisation (PSO) is employed to automatically tune hyperparameters for multiple base classifiers, enabling systematic exploration of the parameter space and improving predictive performance without relying on manual model calibration.

\item A transparent and reproducible evaluation protocol is established using an 80/20 train--test split with stratified cross-validation within the training stage. Feature selection, imbalance handling, and hyperparameter optimisation are performed exclusively on the training data, ensuring that the held-out test set remains unseen during model development and preventing information leakage.

\item Comprehensive benchmarking is conducted across a diverse set of machine learning algorithms and ensemble strategies. Model performance is assessed using multiple metrics, including AUC, F1-score, and Brier score, while bootstrap-based ROC analysis is used to examine the stability and variability of predictive performance across resampled datasets.

\end{enumerate}

The remainder of this paper is organised as follows. Section~\ref{sec:ensem} reviews existing ensemble learning methods and their applications in credit risk modelling. Section~\ref{sec:theoretical_foundation} presents the proposed methodology, including greedy weighting and the BlendNet meta-learner. Section~\ref{sec:data} describes the dataset and preprocessing steps, while Section~\ref{sec:prep} covers feature selection and class imbalance handling. Section~\ref{sec:eva_metrics} introduces the evaluation metrics used. Section~\ref{sec:Results&Intr} reports the results and provides interpretations, and Section~\ref{sec:conc} concludes the study by highlighting contributions and potential implications.

\section{A Review of Ensemble Learning Methods} \label{sec:ensem}

Ensemble learning methods have become central to modern machine learning due to their ability to combine multiple predictive models to improve accuracy, robustness, and generalisation. 
In credit risk modelling, ensemble techniques are particularly valuable because borrower behaviour is heterogeneous and financial variables often exhibit nonlinear interactions that are difficult for single models to capture. 

However, the effectiveness of an ensemble depends critically on how individual learners are aggregated and weighted. In financial environments characterised by evolving borrower profiles, macroeconomic shifts, and class imbalance, static aggregation strategies may become suboptimal. 
This section therefore, reviews the three principal ensemble paradigms--bagging, boosting, and stacking, with emphasis on their practical implications for loan default prediction. 
By identifying their structural limitations in terms of weighting strategies, sequential dependencies, and interpretability, the review provides the conceptual motivation for the proposed GWE framework.

\subsection{Bagging: Bootstrap Aggregation} \label{sec:bagg}

Bootstrap aggregation (bagging), introduced by \citet{breiman1996bagging}, is a foundational ensemble technique that builds multiple versions of a predictor and aggregates their outputs to improve stability and accuracy. The key idea is to generate bootstrap samples from the training data, train independent base models on each sample, and combine their predictions, either through averaging (for regression) or majority voting (for classification). Mathematically, if \( f_1, f_2, \ldots, f_B \) are the models trained on \( B \) bootstrap samples, the final prediction for classification is given by Equation~\eqref{eq:bag_clas}:  

\begin{equation}\label{eq:bag_clas}
\hat{y} = \text{mode}\{f_1(x), f_2(x), \ldots, f_B(x)\}.
\end{equation} 

\noindent Bagging is particularly effective in reducing variance, making it suitable for high-variance models such as decision trees. Random forests, one of the most widely used ensemble methods, extend bagging by introducing feature subsetting to decorrelate trees \citep{breiman2001random}.  

Recent studies have expanded bagging to address emerging challenges in machine learning, particularly in financial risk prediction. For example, \citet{bifet2020adaptive} proposed adaptive online bagging methods to handle streaming and evolving data, enabling models to adapt to concept drift where the statistical relationship between features and labels changes over time. 
Similarly, \citet{Ditzler2015,Gama2014} demonstrated that static ensembles degrade under data drift, especially in non-stationary environments such as credit risk, where borrower profiles and macroeconomic conditions evolve continuously.  

Another limitation of traditional bagging is the assignment of equal weights to all base learners, implicitly assuming homogeneous performance across models. However, in many real-world scenarios, some models may consistently outperform others under specific data regimes.
To address this, recent research has explored weighted bagging strategies, where individual learners receive weights based on criteria such as accuracy, diversity, or recency of performance \citep{He2022}. 
For instance, \citet{Runchi2023} introduced a performance-weighted bagging framework for financial credit scoring that dynamically adjusted learner weights over time, achieving improved robustness under class imbalance and regime shifts. 
Moreover, bagging variants incorporating cost-sensitive learning have emerged to tackle the pervasive issue of class imbalance in credit default datasets.
Approaches such as cost-sensitive bagging \citep{Yotsawat2021} combine resampling techniques with bagging ensembles to penalise misclassifications of minority classes more heavily, thereby enhancing recall for high-risk borrowers while maintaining overall accuracy.
From the perspective of credit risk prediction, the reliance on fixed or heuristically determined weighting strategies can be problematic because borrower characteristics and macroeconomic conditions evolve over time. 
Consequently, the predictive relevance of individual models may vary across different data regimes. 
These limitations highlight the need for ensemble mechanisms that can dynamically reallocate model contributions based on empirical predictive performance.

Despite these advances, existing weighted and adaptive bagging approaches often rely on heuristic or manually tuned weighting schemes, lacking a fully data-driven, greedy optimisation mechanism that continuously prioritises the best-performing models in dynamic environments.
This gap motivates the proposed GWE framework, which extends the bagging philosophy by introducing an empirically driven, adaptive weighting strategy to handle evolving data distributions, class imbalance, and heterogeneous model performance in loan default prediction.

\subsection{Boosting: Sequential Ensemble Learning}\label{sec:boost_seq}

Boosting constructs an ensemble sequentially by training each new learner to correct the errors of its predecessors. A seminal example is AdaBoost \cite{freund1997decision}, which iteratively assigns higher weights to misclassified instances, compelling subsequent models to focus on harder examples. The final prediction combines the weighted contributions of all learners as expressed in Equation~\eqref{eq:boost}:  

\begin{equation}\label{eq:boost}
\hat{y} = \text{sign} \left( \sum_{t=1}^{T} \alpha_t f_t(x) \right),
\end{equation}  

\noindent where \( \alpha_t \) denotes the weight of the \( t \)-th learner, determined by its classification error.
Gradient Boosting (GB) \cite{friedman2001greedy} generalises this principle by optimising arbitrary differentiable loss functions using gradient descent, leading to state-of-the-art implementations such as XGBoost \cite{chen2016xgboost} and Light Gradient Boosting Machine (LightGBM) \cite{ke2017lightgbm}. 
These algorithms have achieved remarkable success in structured data domains, including credit risk prediction, due to their capacity to capture complex nonlinear relationships \cite{Machado2022}.  

In the context of credit risk modelling, boosting algorithms have been adapted to address distinct challenges such as class imbalance, evolving data distributions, and real-time prediction needs. 
For example, \citet{liu2022focal} proposed a focal-aware, cost-sensitive boosting approach for credit scoring, where the loss function is modified to assign higher penalties to misclassified minority-class defaults.
This design enhances the model's ability to detect high-risk loans under severe class imbalance, improving recall without disproportionately sacrificing overall accuracy. 
Similarly, \citet{chang2024machine} developed a framework which applies boosting (among other techniques) on recently collected credit card customer data, demonstrating that adaptive models can improve performance when new borrower information becomes available, thereby alleviating the need for full retraining. 
In a more recent study, \citet{langat2024hybrid} proposed a hybrid XGBoost model for credit risk modelling under changing economic conditions, where model performance is enhanced by adapting to fluctuations such as inflationary pressures. This approach improves robustness compared to conventional boosting methods; however, most existing techniques, including this one, typically address challenges such as imbalance, noise, or drift in isolation. 
This highlights the need for ensemble frameworks capable of simultaneously tackling multiple complexities inherent in credit risk prediction.
In credit scoring applications such as loan default prediction, these limitations become particularly important because datasets often contain evolving borrower behaviour, macroeconomic shocks, and strong class imbalance. 
Under such conditions, boosting methods that primarily focus on instance-level reweighting may not fully capture the changing relative strengths of heterogeneous predictive models.

Despite these advances, existing boosting approaches exhibit several limitations when applied to real-world credit risk prediction. Many cost-sensitive or focal-loss boosting methods improve minority-class detection but neglect the impact of evolving borrower characteristics and macroeconomic conditions on model stability. Conversely, adaptive or hybrid boosting frameworks address concept drift or incremental learning but often assume fixed loss functions or manually tuned weighting schemes, limiting their ability to respond dynamically to simultaneous shifts in data distribution, class imbalance, and noise. Moreover, most boosting variants rely on sequential, additive weight updates rather than performance-driven reallocation across heterogeneous learners, which may underutilise models offering complementary predictive strengths. These gaps underscore the need for ensemble strategies, such as the proposed GWE, that integrate dynamic weighting, drift adaptation, and imbalance handling within a unified, empirically driven framework capable of maintaining accuracy, robustness, and interpretability under changing financial environments.

\subsection{Stacking: Meta-Ensemble Learning} \label{sec:stacking}

Although stacking can achieve strong predictive performance, its application in credit risk modelling raises challenges related to transparency, adaptability, and model dependence. Formally, if \( h_1, h_2, \ldots, h_M \) denote the base models and \( g \) represents the meta-learner, the final prediction is expressed as Equation~\eqref{eq:stack}:  

\begin{equation}\label{eq:stack}
\hat{y} = g(h_1(x), h_2(x), \ldots, h_M(x)).
\end{equation} 

\noindent Since its introduction by \citet{wolpert1992stacked}, stacking has become a widely used meta-ensemble technique in credit risk prediction, particularly with the advent of robust meta-learners such as gradient boosting machines, penalised regressions, and neural networks \cite{yin2023stacking}. These modern variants exploit the predictive diversity of heterogeneous base models, often achieving superior accuracy compared to single models or homogeneous ensembles.

Despite these advantages, several limitations persist in existing stacking approaches. First, most meta-learners implicitly infer weights for base models through their internal optimisation process, offering limited transparency and potentially overemphasising weak or redundant models.
In financial domains, where interpretability and regulatory compliance are critical, such opacity hinders practical adoption. Second, conventional stacking typically assumes that base models exhibit complementary error patterns; however, empirical studies reveal that high-capacity learners often produce correlated predictions, diminishing the benefits of aggregation \cite{yin2023stacking}. Third, standard meta-learners lack mechanisms to adaptively adjust model contributions under evolving data distributions or market regimes, a limitation particularly acute in credit risk modelling where borrower profiles and macroeconomic conditions shift over time \cite{langat2024hybrid}.  

Recent work has sought to address these gaps through adaptive and hybrid stacking strategies. For example, \citet{yin2023stacking} proposed a hierarchical stacking ensemble for peer-to-peer lending, integrating feature selection and clustering-based model grouping to enhance diversity before meta-learning. Similarly, \citet{chang2024machine} demonstrated that periodically retraining stacking ensembles on newly available credit data improves robustness under concept drift. More innovative approaches employ attention mechanisms or probabilistic weighting to dynamically modulate base-model influence in response to performance fluctuations \cite{yang2023attention}. However, these methods typically optimise weights within the meta-learner's training phase rather than continuously updating them in real-time or across changing environments.

This limitation directly motivates the GWE proposed in this study. Unlike conventional stacking, which delegates weight assignment entirely to the meta-learner, GWE introduces a performance-driven, iterative weighting mechanism operating alongside the ensemble.
By continuously monitoring empirical model accuracy over rolling time windows, GWE adaptively reallocates weights to prioritise high-performing models while down-weighting weaker or redundant learners. 
This hybridisation of stacking flexibility with dynamic weighting not only improves predictive stability under non-stationary conditions but also enhances interpretability by explicitly revealing model contributions at each stage. Consequently, GWE bridges the gap between static stacking ensembles and the adaptive, transparent architectures increasingly demanded in modern credit risk prediction.

\subsection{Contemporary Approaches in Tabular and Financial Data Modelling} \label{sec:tabular}

It is important to distinguish between modelling paradigms designed for sequential financial time-series data and those suitable for cross-sectional credit scoring tasks. 
The LendingClub dataset used in this study consists of individual loan application records with borrower attributes measured at the time of loan issuance. 
Consequently, the prediction task is formulated as a tabular binary classification problem rather than a temporally ordered forecasting problem. 
While sequential architectures such as recurrent neural networks (RNNs) and Long Short-Term Memory (LSTM) models are widely used for financial time-series forecasting, their applicability is limited when explicit temporal sequences are not present in the data. 
For structured tabular datasets of this nature, ensemble machine learning methods remain among the most effective and widely adopted modelling strategies.

Recent years have witnessed a growing interest in advanced machine learning architectures for tabular and financial datasets, driven by the increasing availability of high-dimensional borrower information and the need for accurate credit risk assessment. Among these, transformer-based models have attracted considerable attention due to their ability to capture complex feature dependencies through attention mechanisms. 
Models such as TabNet \cite{arik2021tabnet}, SAINT \cite{somepalli2021saint}, and TabTransformer \cite{huang2020tabtransformer} adapt ideas from natural language processing to structured data, employing self-attention layers to learn context-aware feature representations.
Empirical studies have demonstrated their competitive performance across classification and regression tasks in domains ranging from credit default prediction to fraud detection \cite{gorishniy2021revisiting}. Nevertheless, these architectures are often criticised for their computational cost, sensitivity to hyperparameter tuning, and limited interpretability factors that constrain their adoption in highly regulated financial settings where transparency and operational efficiency remain paramount \cite{chen2023interpretable}.

Parallel to the development of attention-based models, deep ensemble methods have emerged as another prominent direction in financial risk prediction. By aggregating multiple neural networks or gradient boosting models, deep ensembles aim to improve predictive stability and quantify model uncertainty \cite{lakshminarayanan2017simple}. In the credit risk domain, \citet{yin2023stacking} introduced a stacking ensemble combining gradient boosting, support vector machines, and logistic regression, reporting significant gains in prediction accuracy for peer-to-peer lending platforms. Similarly, \citet{chang2024machine} employed hybrid ensembles incorporating both tree-based learners and neural architectures to capture nonlinear interactions in evolving credit datasets. While these approaches often achieve state-of-the-art accuracy, they frequently require extensive model training pipelines, involve non-trivial feature engineering, and lack mechanisms for dynamic adaptation when borrower profiles or economic conditions shift over time.

Moreover, both transformer-based and deep ensemble approaches frequently prioritise predictive performance at the expense of interpretability and adaptability.
As highlighted by \citet{langat2024hybrid}, models optimised solely for accuracy risk underperform when exposed to class imbalance, regime shifts, or noisy financial records, realities commonly encountered in credit scoring applications. These challenges motivate the development of frameworks capable of balancing accuracy, efficiency, and transparency while remaining robust to changing data environments.

The proposed GWE framework directly addresses these gaps by integrating a diverse collection of classical machine learning models through a performance-driven, dynamically updated weighting strategy. Unlike transformer-based methods that rely on deep attention layers or stacking ensembles with static meta-learners, GWE continuously monitors the empirical performance of each base model and reallocates weights to prioritise strong learners under current data conditions. This design reduces computational complexity, improves interpretability by explicitly revealing model contributions, and enhances adaptability to class imbalance and concept drift. Consequently, GWE offers a practical and scalable alternative for credit risk modelling, complementing recent advances while overcoming critical limitations in existing ensemble and attention-based approaches.

Table~\ref{tab:ensemble_comparison} summarises the conceptual differences between traditional ensemble paradigms and highlights the limitations that motivate the proposed Greedy Weighting Ensemble framework.

\begin{table*}[ht]
\centering
\caption{Comparison of ensemble paradigms and their limitations in credit risk modelling}
\label{tab:ensemble_comparison}
\begin{tabular}{p{3cm} p{3.5cm} p{4cm} p{4cm}}
\hline
\textbf{Paradigm} & \textbf{Core Principle} & \textbf{Strengths} & \textbf{Key Limitations for Credit Risk Prediction} \\
\hline

Bagging &
Independent models trained on bootstrap samples and aggregated through voting or averaging &
Reduces variance and improves model stability &
Typically assigns equal or heuristic weights to base learners, limiting adaptability under concept drift and evolving borrower behaviour \\

Boosting &
Sequential training where each learner focuses on correcting previous errors &
Strong predictive accuracy and ability to capture nonlinear patterns &
Sequential dependence may amplify noise or instability; primarily reweights observations rather than dynamically reallocating model importance \\

Stacking &
Meta-learner combines predictions from heterogeneous base models &
High predictive flexibility and ability to exploit model diversity &
Model contributions are often implicit and opaque; limited mechanisms for adapting model influence under changing data regimes \\

\hline

Proposed GWE &
Performance-driven greedy optimisation of ensemble weights &
Adaptive model weighting, improved interpretability, and robustness to evolving data distributions &
Designed to overcome static weighting, sequential dependence, and opacity limitations present in traditional ensembles \\

\hline
\end{tabular}
\end{table*}

\section{Greedy Optimisation for Model Weighting in Ensemble Learning}\label{sec:theoretical_foundation}

Greedy optimisation is an efficient algorithmic strategy that iteratively makes locally optimal choices to pursue a global optimum. In ensemble learning, it is beneficial for model weighting to assign different importance to base models based on their predictive performance. By iteratively adjusting weights to minimise a loss function, greedy optimisation balances individual contributions within the ensemble.
Recent studies \citep{alocen2024greedy, wang2024martingale, zhao2024iterative, mao2011greedy} have demonstrated its effectiveness in improving ensemble accuracy and diversity. For instance, \citet{mao2011greedy} introduced the Matching Pursuit Optimisation Ensemble Classifiers (MPOEC), which uses a greedy algorithm to iteratively refine model weights, enhancing both individual accuracy and classifier diversity. Such approaches consistently outperform traditional ensemble strategies, particularly in high-dimensional data scenarios.

\subsection{Base Model Selection} \label{sec:base_model}

The first critical step in the greedy weighting framework is the selection of base models. Base models are individual learning algorithms trained on the same dataset, whose predictions are subsequently combined within an ensemble. In this study, a diverse set of classifiers is employed, including Gradient Boosting (GB), Multi-Layer Perceptron (MLP), Support Vector Machines (SVM), k-Nearest Neighbors (KNN), Logistic Regression (LR), and Extra Trees. 
This diverse selection ensures that the ensemble benefits from the complementary strengths of different learning paradigms. For example, GB effectively captures non-linear relationships in structured, tabular data, while MLPs are capable of modelling complex, high-dimensional feature interactions. Similarly, KNN can identify local decision boundaries, whereas linear models like LR offer interpretability and robustness. 
The diversity among these base models--in terms of their underlying structures, learning mechanisms, and inductive biases is essential to the ensemble's overall predictive performance.
Combining models with varying strengths and weaknesses helps mitigate the risk of overfitting to specific data characteristics and allows the ensemble to correct individual model errors.
Additionally, each base model is further refined through customised hyperparameter tuning, where these optimised settings critically shape the model's generalisation ability by managing the balance between bias and variance.

\subsection{Optimising Hyperparameters with Particle Swarm Optimisation}\label{sec:pso}

Optimising hyperparameters is a critical task in ML, as it has a direct effect on a model’s performance. 
Common methods for uncovering optimal hyperparameters are Grid-search, Random-search, and Bayesian optimisation.
Despite their popularity, they each come with certain drawbacks, especially when applied to high-dimensional spaces or extensive datasets.
Grid-search involves a thorough search over a predetermined set of hyperparameter combinations and can be quite demanding in terms of computational resources, particularly as the number of hyperparameters escalates. 
This exhaustive method encounters the curse of dimensionality \cite{altman2018curse,bengio2005curse,bellman1961adaptive}, meaning the evaluations needed rise exponentially with the increasing dimensionality of the hyperparameter space.
For instance, a model with five hyperparameters, each having ten possible values, necessitates \(10^5\) evaluations, rendering it impractical for complex models or large datasets \cite{bergstra2012random}.  
Random-search alleviates this inefficiency by sampling hyperparameters randomly from predefined distributions.
Though more scalable than Grid-search, its unguided exploration lacks mechanisms to balance exploration (searching new regions) and exploitation (refining known promising regions), often resulting in suboptimal convergence \cite{bardenet2013collaborative}. 
Bayesian optimisation addresses these shortcomings by constructing probabilistic surrogate models (e.g., Gaussian processes) to predict optimal hyperparameters iteratively. While efficient in low-dimensional spaces, its performance degrades in noisy, discontinuous, or high-dimensional domains due to reliance on smoothness assumptions \cite{snoek2012practical}.
Additionally, Bayesian methods require careful tuning of acquisition functions (e.g., Expected Improvement) and struggle with discrete parameters, limiting their versatility \cite{shahriari2016unbounded}.  

In this study, Particle Swarm Optimisation (PSO) is adopted as a practical population-based search strategy for hyperparameter tuning. 
The objective is not to compare optimisation algorithms but to obtain well-calibrated base learners before ensemble weighting is applied. 
PSO provides an efficient mechanism for exploring heterogeneous hyperparameter spaces without relying on strong smoothness assumptions about the optimisation landscape. 
PSO is a population-based optimisation method that draws inspiration from the social behaviour of birds flocking or fish schooling \cite{kennedy1995particle,eberhart1995particle}. 
It starts with a group of potential solutions (particles), each one embodying a candidate set of hyperparameters.
These particles are initially placed randomly and then moved iteratively within the hyperparameter space, their movement guided by the evaluation of fitness, which is typically linked to the model’s performance. 
The particles adjust their position based on their own previous best result and the best solution found by the entire swarm.
This collective intelligence permits the swarm to hone in on optimal hyperparameters without the exhaustive and costly exploration characteristic of Grid-search or the randomness inherent in Random-search. 
Notable benefits of PSO include the following:

\begin{enumerate}[i.]
  \item PSO strikes an effective balance between exploring the search space for new, potentially better solutions and exploiting known good solutions by refining the search around promising areas.
  
  \item Unlike Grid-search, which evaluates hyperparameter combinations sequentially, PSO evaluates multiple solutions simultaneously. This parallel evaluation leads to faster convergence and more efficient exploration.
  
  \item PSO is computationally more efficient than exhaustive methods, such as Grid-search, especially in high-dimensional hyperparameter spaces. The swarm-based approach reduces the overall computational burden by focusing on areas that are more likely to yield better results.
\end{enumerate}

\noindent In PSO, the optimisation problem is framed as finding the best possible hyperparameters in a predefined search space. The position of each particle represents a possible set of hyperparameters, and the particle’s velocity determines how it moves through the search space. The algorithm is iterative, with particles adjusting their positions based on both their individual experiences and the experiences of the entire swarm.
Let the position of the \(i\)-th particle be represented as a vector presented in Equation~\eqref{eq:par}:

\begin{equation}\label{eq:par}
\mathbf{p}_i = [p_{i1}, p_{i2}, \dots, p_{id}],
\end{equation}

\noindent  where \(p_{ij}\) is the value of the \(j\)-th hyperparameter for the \(i\)-th particle. The velocity of each particle is also represented as a vector depicted in Equation~\eqref{eq:vel}:

\begin{equation}\label{eq:vel}
\mathbf{v}_i = [v_{i1}, v_{i2}, \dots, v_{id}].
\end{equation}

\noindent  The velocity update rule governs how particles move in the hyperparameter space and is given by Equation~\eqref{eq:ad_3}:

\begin{align}
\label{eq:ad_3}
v_{ij}(t+1) &= w \cdot v_{ij}(t) + c_1 \cdot \text{rand}_1 \cdot (p_{ij}^{best} - p_{ij}) \notag \\
&\quad + c_2 \cdot \text{rand}_2 \cdot (g_{j}^{best} - p_{ij}),
\end{align}

\noindent  where \(v_{ij}(t)\) is the velocity of the \(i\)-th particle for the \(j\)-th hyperparameter at time step \(t\),
\(w\) is the inertia weight controlling the previous velocity’s influence,
\(c_1\) and \(c_2\) are cognitive and social learning factors, respectively,
\(\text{rand}_1\) and \(\text{rand}_2\) are random numbers in the range \([0, 1]\),
\(p_{ij}^{best}\) is the best-known position of the \(i\)-th particle for the \(j\)-th hyperparameter,
\(g_{j}^{best}\) is the best-known global position for the \(j\)-th hyperparameter.

\noindent  The position update rule determines how each particle moves based on its velocity (refer to Equation~\eqref{eq:par_2}):

\begin{equation}\label{eq:par_2}
p_{ij}(t+1) = p_{ij}(t) + v_{ij}(t+1).
\end{equation}

\noindent  The velocity is then clipped to ensure it remains within a reasonable range, preventing the particles from making excessively large moves through the search space as defined in  Equation~\eqref{eq:vel_2}:

\begin{equation}\label{eq:vel_2}
v_{ij} = \min(\max(v_{ij}, -v_{\text{max}}), v_{\text{max}}),
\end{equation}

\noindent  where \(v_{\text{max}}\) is the maximum allowed velocity.
Each particle’s fitness is evaluated based on a fitness function \(f(\mathbf{p}_i)\), which in the context of hyperparameter optimisation is typically the validation score of a machine learning model. The goal is to find the hyperparameter set \(\mathbf{p}_i\) that maximises (or minimises, depending on the problem) the fitness function. The best positions (\(p_{ij}^{best}\)) are updated whenever a particle achieves a better fitness score, and the global best (\(g_j^{best}\)) is updated accordingly. 
This process continues iteratively, with the particles exploring and exploiting the search space, converging toward an optimal set of hyperparameters over time.
Below is the pseudocode (refer to Algorithm~\ref{al:pso-hyperparameter}) for the PSO algorithm used for hyperparameter tuning:

\begin{algorithm}[H]
\caption{PSO for hyperparameter tuning}\label{al:pso-hyperparameter}
\begin{algorithmic}[1]
\Require Hyperparameter search space $\mathcal{S}$, fitness function $f(\cdot)$
\Ensure Optimal hyperparameters $\mathbf{p}_{opt}$

\State Initialize population $\mathbf{P} = \{\mathbf{p}_1, \mathbf{p}_2, \dots, \mathbf{p}_N\}$, where each particle $\mathbf{p}_i$ is a random hyperparameter set
\State Initialize particle velocities $\mathbf{V} = \{\mathbf{v}_1, \mathbf{v}_2, \dots, \mathbf{v}_N\}$
\State Initialize personal best positions $\mathbf{p}_i^{best} = \mathbf{p}_i$ and global best position $\mathbf{g}^{best}$
\State Set parameters: cognitive factor $c_1$, social factor $c_2$, inertia weight $w$, maximum velocity $v_{max}$

\For{iteration $t=1$ to $T$}
    \For{each particle $\mathbf{p}_i$}
        \State Evaluate fitness $f(\mathbf{p}_i)$
        \If{$f(\mathbf{p}_i) > f(\mathbf{p}_i^{best})$}
            \State Update personal best: $\mathbf{p}_i^{best} = \mathbf{p}_i$
        \EndIf
        \If{$f(\mathbf{p}_i^{best}) > f(\mathbf{g}^{best})$}
            \State Update global best: $\mathbf{g}^{best} = \mathbf{p}_i^{best}$
        \EndIf
    \EndFor
    \For{each particle $\mathbf{p}_i$}
        \State Update velocity: 
        \[
        \mathbf{v}_i = w \cdot \mathbf{v}_i + c_1 \cdot \text{rand}_1 \cdot (\mathbf{p}_i^{best} - \mathbf{p}_i) + c_2 \cdot \text{rand}_2 \cdot (\mathbf{g}^{best} - \mathbf{p}_i)
        \]
        \State Update position: $\mathbf{p}_i = \mathbf{p}_i + \mathbf{v}_i$
        \State Clip velocity: $$\mathbf{v}_i = \min(\max(\mathbf{v}_i, -v_{max}), v_{max})$$
    \EndFor
\EndFor

\State \Return $\mathbf{g}^{best}$ as the optimal hyperparameter set
\end{algorithmic}
\end{algorithm}

\noindent For each base model, the hyperparameter space is carefully defined to optimise critical parameters that influence model performance.
Table \ref{tab:pso_hyperparams} summarises the key hyperparameters for each base model and their optimised values after applying PSO.

\begin{table*}[htbp]
\centering
\caption{Comparison of hyperparameter spaces and optimal values for different base models tuned using PSO}
\label{tab:pso_hyperparams}
\begin{tabular}{|l|p{6cm}|p{6cm}|}
\hline
\textbf{Base model} & \textbf{Defined hyperparameter space} & \textbf{Best optimal parameter (via PSO)} \\
\hline
GB & 
n\_estimators: (50, 200), max\_depth: (3, 10), learning\_rate: (0.01, 0.1) & 
n\_estimators: 150, max\_depth: 5, learning\_rate: 0.05 \\
\hline
MLP & 
hidden\_layer\_sizes: (50, 200), alpha: (0.0001, 0.01) & 
hidden\_layer\_sizes: 150, alpha: 0.005 \\
\hline
SVM & 
C: (0.1, 10), gamma: (0.001, 0.1) & 
C: 3, gamma: 0.05 \\
\hline
KNN & 
n\_neighbors: (3, 20) & 
n\_neighbors: 10 \\
\hline
LR & 
C: (0.1, 10) & 
C: 1.5 \\
\hline
Extra Trees & 
n\_estimators: (50, 200), max\_depth: (10, 20), min\_samples\_split: (2, 10) & 
n\_estimators: 120, max\_depth: 15, min\_samples\_split: 4 \\
\hline
\end{tabular}
\end{table*}

\subsection{Mathematical Formulation of Greedy Weighting}

The greedy weighting method is formulated as an iterative process that minimises the classification loss. For each weight \(w_i\), the algorithm searches a predefined range of possible values and updates the weight based on the improvement in the ensemble's accuracy. The greedy process is expressed by Equation~\eqref{eq:greedy-pro}:


\begin{equation}\label{eq:greedy-pro}
w^{(k+1)}_i \;=\; \arg\min_{w_i \in [0,1]} \; 
\Bigg\{ \; \frac{1}{m}\sum_{j=1}^{m} \big(\hat{y}_j(w^{(k)}_1,\dots,w^{(k)}_{i-1},w_i,w^{(k)}_{i+1},\dots,w^{(k)}_n)-y_j\big)^2 \;+\; \lambda \sum_{t=1}^n w_t^2 \Bigg\},
\end{equation}

\noindent where \(w_i^{(k)}\) is the weight for the \(i\)-th model at the \(k\)-th iteration, \(m\) is the number of validation samples, \(y_j\) is the true label for the \(j\)-th sample, \(\hat{y}_j(w_1^{(k)}, \ldots, w_n^{(k)})\) is the ensemble prediction for the \(j\)-th sample under the current weights, and \(\lambda\) is a regularisation parameter. The iterative process continues until convergence, defined as the point where no further significant improvement in ensemble performance is observed.
Although the final prediction task is binary classification, the ensemble optimisation operates on probabilistic predictions produced by the base models. 
Under this formulation, the squared-error objective corresponds to the Brier loss, a proper scoring rule widely used for evaluating probabilistic forecasts in classification problems. 
Minimising this objective therefore encourages well-calibrated probability estimates while simultaneously determining the optimal ensemble weights.

The greedy optimisation method is further supplemented with regularisation to prevent overfitting, which could arise from excessive reliance on a small subset of base models. 
To stabilise the ensemble weighting process, we introduce a quadratic penalty on the weight vector, resulting in the regularised objective function given in Equation~\eqref{eq:regularisation}.

\begin{equation}\label{eq:regularisation}
L(\mathbf{w}) \;=\; \frac{1}{m}\sum_{j=1}^{m}\big(\hat{y}_j(\mathbf{w})-y_j\big)^2 \;+\; \lambda \sum_{i=1}^n w_i^2,
\end{equation}

\noindent where \(\lambda > 0\) is the regularisation parameter controlling the trade-off between prediction error and the magnitude of the ensemble weights. 
The quadratic penalty term \(\lambda \sum_{i=1}^n w_i^2\) discourages extreme weight assignments by shrinking large coefficients, thereby preventing the ensemble from becoming dominated by a single model. 
This form of regularisation improves numerical stability and promotes smoother weight distributions across the ensemble.

In our implementation, we additionally impose non-negativity and convexity constraints on the weights,
\[
w_i \ge 0 \quad \forall i,\qquad \sum_{i=1}^n w_i = 1,
\]
ensuring that the final prediction remains an interpretable convex combination of the base learners. 
These constraints prevent pathological solutions in which weights become arbitrarily large or oscillate during optimisation.
From a practical perspective, these mechanisms also mitigate the potential effect of short-term fluctuations when model performance is evaluated over a rolling validation window. 
Financial datasets may occasionally contain transient anomalies or noisy observations that temporarily affect the predictive performance of individual models. 
The quadratic regularisation term discourages abrupt changes in the weight vector, while the convexity constraint ensures that model contributions remain bounded and balanced. 
These constraints stabilise the greedy weighting procedure and reduce sensitivity to short-lived performance fluctuations.

The pseudocode in Algorithm~\ref{al:greedy-weighting-regularized} describes a greedy weighting with regularisation algorithm for optimising the weights of base models in an ensemble to minimise prediction loss.

\begin{algorithm}[H]
\caption{Greedy weighting with regularisation for optimised model ensemble}
\label{al:greedy-weighting-regularized}
\begin{algorithmic}[1]
\Require Set of base models $\mathcal{M} = \{M_1, M_2, \dots, M_n\}$, validation dataset $(X_{val}, y_{val})$, loss function $L(\cdot, \cdot)$, regularisation parameter $\lambda$
\Ensure Optimized weights $\mathbf{w} = \{w_1, w_2, \dots, w_n\}$

\State Initialize weights $\mathbf{w} = \{w_1 = 1/n, w_2 = 1/n, \dots, w_n = 1/n\}$ \Comment{Equal weights initially}
\State Compute initial ensemble prediction: $\hat{y}_{ensemble} = \sum_{i=1}^n w_i M_i(X_{val})$
\State Evaluate initial loss with regularisation: $L_{ensemble} = L(y_{val}, \hat{y}_{ensemble}) + \lambda \sum_{i=1}^n w_i^2$ \Comment{Penalize large weights}

\For{$i = 1$ to $n$}
    \State Set $\Delta w = \delta$ \Comment{Small weight increment}
    \State Update weight of model $M_i$: $w_i = w_i + \Delta w$
    \State Normalize weights: $w_i = w_i / \sum_{j=1}^n w_j$
    \State Compute new ensemble prediction: $\hat{y}_{new} = \sum_{j=1}^n w_j M_j(X_{val})$
    \State Evaluate new loss with regularisation: $L_{new} = L(y_{val}, \hat{y}_{new}) + \lambda \sum_{j=1}^n w_j^2$

    \If{$L_{new} < L_{ensemble}$}
        \State Accept new weight: $L_{ensemble} = L_{new}$
    \Else
        \State Revert weight: $w_i = w_i - \Delta w$
        \State Normalize weights: $w_i = w_i / \sum_{j=1}^n w_j$
    \EndIf
\EndFor

\State \Return Optimized weights $\mathbf{w}$
\end{algorithmic}
\end{algorithm}

\noindent Once the base models (SVM, Extra Trees, GB, MLP, KNN, LR) are optimised through hyperparameter tuning, their predictions are combined using a customised greedy-weighted ensemble framework. This approach assigns optimised, data-driven weights to each base model based on their predictive performance, rather than relying on standard homogeneous weighting ensemble techniques adopted in Bagging, Boosting, or Stacking.
By employing the greedy optimisation strategy outlined in Section~\ref{sec:ensem}, the ensemble iteratively adjusts model contributions to minimise classification error, resulting in a more adaptive, efficient, and performance-focused aggregation of base learners.

Bagging, or Bootstrap Aggregating, involves training multiple instances of base models on different bootstrapped subsets of the dataset and averaging their predictions. The optimised weights \( w_i \) are used to adjust the contribution of each base model to the final aggregated prediction. Let \( f_i(x) \) be the prediction of the \(i\)-th base model. 
The final bagging prediction is computed as Equation~\eqref{eq:gbagging}:

\begin{equation} \label{eq:gbagging}
\hat{y}_{\text{bagging}} = \sum_{i=1}^{n} w_i f_i(x),
\end{equation}

\noindent where the greedy-optimised weights \(w_i\) ensure that stronger base models contribute more significantly to the final decision, while weaker models are down-weighted.

In the case of Boosting, models are trained sequentially, with each new model focusing on correcting the errors made by its predecessors. The optimised weights \( w_i \) from the greedy-weighting process can be incorporated into a modified loss function, guiding how subsequent models adjust for previous errors. At each boosting iteration \( t \), a model \( f_t(x) \) is trained with an updated weight function defined in Equation~\eqref{eq:gboost}:

\begin{equation} \label{eq:gboost}
w_i^{(t+1)} = w_i^{(t)} e^{-\alpha_t (y_i f_t(x))},
\end{equation}

\noindent where \( \alpha_t \) is the learning rate, and misclassified samples receive larger updates in weight. By using only well-tuned models, greedy-weighting improves generalisation and reduces overfitting in the boosting process.

Finally, Stacking combines predictions from multiple base models by feeding them into a meta-learner. The optimised base models act as first-level learners, and their predictions serve as inputs to a higher-level model, typically logistic regression or another ensemble approach. Given base model predictions as $h_1(x), h_2(x), \ldots, h_n(x)$; 
the final stacking prediction depicted in Equation~\eqref{eq:gstack} is made by the meta-model \( g(\cdot) \):

\begin{equation} \label{eq:gstack}
\hat{y}_{\text{stacking}} = g\left(w_1 h_1(x), w_2 h_2(x), ..., w_n h_n(x)\right).
\end{equation}

\noindent In this structure, the greedy-optimised weights ensure that models with greater predictive accuracy are prioritised, ultimately enhancing the overall performance of the stacked ensemble.

\subsection{BlendNet: A Neural Network-Based Meta-Learner}\label{sec:blendnet}

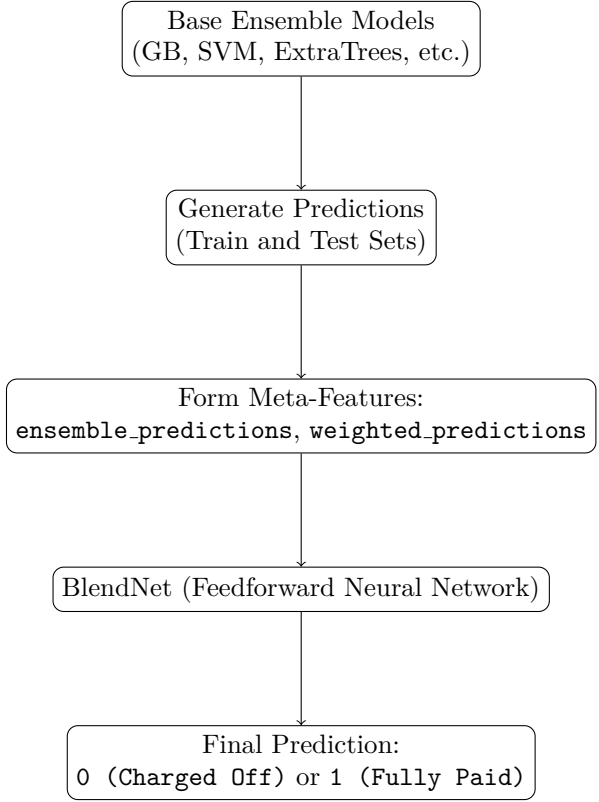
\begin{figure}[H]
\centering
\begin{tikzpicture}[node distance=1.5cm, every node/.style={draw, rounded corners, align=center}]
\node (base) {Base Ensemble Models \\ (GB, SVM, ExtraTrees, etc.)};
\node (preds) [below=of base] {Generate Predictions \\ (Train and Test Sets)};
\node (meta) [below=of preds] {Form Meta-Features: \\ \texttt{ensemble\_predictions}, \texttt{weighted\_predictions}};
\node (blendnet) [below=of meta] {BlendNet (Feedforward Neural Network)};
\node (output) [below=of blendnet] {Final Prediction: \\ \texttt{0 (Charged Off)} or \texttt{1 (Fully Paid)}};

\draw[->] (base) -- (preds);
\draw[->] (preds) -- (meta);
\draw[->] (meta) -- (blendnet);
\draw[->] (blendnet) -- (output);
\end{tikzpicture}
\caption{Flowchart illustrating the BlendNet meta-learning process within the stacked ensemble framework.}
\label{fig:blendnet-flowchart}
\end{figure}

To effectively aggregate predictions from multiple base ensemble models, we introduce a neural network-based meta-learner, BlendNet, operating within a stacked ensemble learning framework. 
This meta-learner serves as the final decision layer by learning from meta-features constructed by combining the predictions of both individually weighted and unweighted ensemble models. 
Through this design, the meta-learner captures higher-order relationships and dependencies among the ensemble outputs, enabling a robust and adaptive final classification. The architecture consists of a fully connected feedforward neural network with three fully connected hidden layers, each consisting of neurons that perform weighted linear transformations followed by nonlinear activation functions \cite{Goodfellow2016deep}.
Its input layer receives a meta-feature matrix comprising predictions from the base models, generated separately for the training and testing datasets. These matrices, which consolidate both ensemble and greedy-weighted predictions, are subsequently processed by the meta-learner to produce the final classification outputs, as illustrated in Fig.~\ref{fig:blendnet-flowchart}.

\noindent The BlendNet model is implemented as a sequence of densely connected layers with ReLU activation functions, followed by batch normalisation and dropout regularisation to mitigate overfitting. The final output layer applies a sigmoid activation function to perform binary classification. This implementation ensures that the meta-learner effectively models complex relationships within the meta-feature space while maintaining generalisation capability. The detailed procedural representation of the BlendNet architecture is provided in Algorithm~\ref{al:blendnet}.

\begin{algorithm}[H]
\caption{BlendNet Meta-Learner for Ensemble Stacking}\label{al:blendnet}
\begin{algorithmic}[1]
\Require Meta-feature matrices $\mathbf{X}_{train}, \mathbf{X}_{test}$, labels $\mathbf{y}_{train}, \mathbf{y}_{test}$, number of epochs $E$, batch size $B$
\Ensure Final predicted labels $\hat{\mathbf{y}}_{test}$

\State Define BlendNet architecture:
\Statex \hspace{1em} Input layer of dimension $d$ (number of meta-features)
\Statex \hspace{1em} Dense layer with 128 neurons, ReLU activation
\Statex \hspace{1em} Batch Normalisation layer
\Statex \hspace{1em} Dropout layer with rate 0.3
\Statex \hspace{1em} Dense layer with 64 neurons, ReLU activation
\Statex \hspace{1em} Dense layer with 32 neurons, ReLU activation
\Statex \hspace{1em} Output layer with 1 neuron, sigmoid activation

\State Compile model using Adam optimiser and binary cross-entropy loss
\State Fit model on $\mathbf{X}_{train}, \mathbf{y}_{train}$ for $E$ epochs with batch size $B$, validating on $(\mathbf{X}_{test}, \mathbf{y}_{test})$
\State Predict probabilities on $\mathbf{X}_{test}$
\State Threshold probabilities at 0.5 to obtain final binary predictions $\hat{\mathbf{y}}_{test}$
\State \Return $\hat{\mathbf{y}}_{test}$
\end{algorithmic}
\end{algorithm}

\noindent BlendNet complements the GWE framework by combining diverse ensemble outputs into a single, optimised prediction. This approach improves the robustness, accuracy, and generalisation capabilities of the overall loan default prediction model.

Although the individual components of the proposed pipeline, greedy weighting, PSO for hyperparameter search, and neural meta-learner stacking, have precedents in the literature, the present work advances the state of practice in three interrelated ways. First, we apply a regularised continuous greedy weighting scheme (softmax parameterisation with an $L_2$ penalty) that yields convex, trackable model weights rather than performing hard selection; this contrasts with matching-pursuit-style selective ensembles that iteratively add or remove classifiers \cite{Mao2011,Jiao2006}. Second, per-model PSO tuning is performed prior to ensemble weighting so that weights reflect each learner's optimised capability rather than untuned performance. 
This decomposition reduces cross-dependence between hyperparameter search and ensemble optimisation \cite{engelbrecht2014particle}.
Third, the BlendNet meta-learner is explicitly fed both weighted and unweighted ensemble outputs alongside base predictions, enabling the meta-learner to correct for residual calibration and capture higher-order dependencies among model outputs. Taken together, these design choices produce a method that emphasises discriminative performance, calibrated probability estimates and model transparency, properties that are especially relevant to credit risk applications.

\section{Dataset Insights and Feature Analysis}\label{sec:data}
\subsection{Dataset Description}\label{sec:description}

The dataset used for this research was sourced from Lending Club\footnote{https://www.lendingclub.com/personal-savings/founder-savings}, a prominent online peer-to-peer lending platform that has gained significant recognition for its role in modernising personal loan transactions. 
Lending Club's business model is structured to connect borrowers seeking personal loans with investors looking for opportunities to earn returns through fixed-income investments. 
Borrowers submit loan applications through the Lending Club website, providing detailed personal and financial information, including such as credit scores, annual income, employment history, and debt-to-income ratios. 
These details are assessed to determine creditworthiness, after which approved loans are listed on the platform.
Investors, both individual and institutional, can then fund these loans by purchasing notes representing portions of the total loan amount. 
Once the loan is fully funded, the approved amount is disbursed to the borrower. 
Borrowers repay their loans in fixed monthly installments comprising both principal and interest, which are subsequently distributed to the investors proportionally. 
Lending Club also features a secondary market, allowing investors to trade existing loans, providing liquidity for those who wish to exit their investment before the loan matures.
For this research, the Lending Club dataset was chosen due to its comprehensiveness and relevance to the problem of loan default prediction. 
The dataset offers detailed information on borrowers and their loan performance, including loan amounts, interest rates, loan grades, annual income, credit scores, and loan status (e.g., fully paid or charged off).
More importantly, the dataset is publicly available\footnote{https://www.kaggle.com/datasets/epsilon22/lending-club-loan-two}, making it accessible for academic research while adhering to ethical and data protection considerations. 

Furthermore, its real-world nature makes it an ideal candidate for evaluating the practical applicability and robustness of predictive models in the financial domain.
The richness of the Lending Club dataset allows for the exploration of critical features relevant to loan default prediction, such as credit scores, income levels, employment lengths, and loan purposes.
The inclusion of both categorical features (e.g., loan grade and loan purpose) and numerical features (e.g., annual income and interest rates) ensures that the data captures a diverse range of factors influencing loan repayment behaviour. 
This diversity also poses challenges, such as handling mixed data types, feature engineering, and addressing missing or inconsistent values--issues that are thoroughly addressed in the pre-processing phase of this study.
Another key factor influencing the choice of this dataset is its representation of real-world class imbalance, where the number of borrowers who repay loans far exceeds those who default.
This mirrors typical financial datasets, where defaults are relatively rare events but have significant implications for financial institutions. 
The imbalance presents an excellent opportunity to demonstrate the efficacy of the proposed greedy weighting ensemble framework in handling such challenges.
The dataset also provides historical trends in loan performance, enabling the exploration of temporal dynamics in borrower behaviour. This temporal aspect is particularly valuable for stress-testing the model under varying economic conditions and understanding the influence of macroeconomic factors on loan defaults.

\begin{table}[htbp]
\centering 
\caption{Description of loan variables.} 
\label{table:var}
\begin{tabular}{l|l p{10cm}}
\toprule
{} &           Items &                                        Description \\
\midrule
0  &             loan\_amnt & The specified loan amount requested by the borrower \\
1  &                  term &  The quantity of installments for the loan  \\
2  &              int\_rate &      Loan's interest rate\\
3  &           installment &  The borrower's monthly payment obligation \\
4  &                 grade &   Loan grade assigned by LC \\
5  &             sub\_grade &   Subgrade designation for LC-assigned loans \\
6  &             emp\_title &  The Borrower's provided job title  \\
7  &            emp\_length &  Duration of employment in years \\
8  &        home\_ownership &  The borrower's provision of homeownership status \\
9  &            annual\_inc & The borrower's self-declared yearly income submitted during the registration process \\
10 &   verification\_status &  Specifies whether income has been validated by LC, remains unverified, or if the source of income has undergone verification \\
11 &               issue\_d &       The month during which the loan was provided \\
12 &           loan\_status &          Loan Status Update \\
13 &               purpose &  The borrower designates a category for the loan application \\
14 &                 title &            The title of the loan as supplied by the borrower \\
15 &              zip\_code &  The initial three digits of the postal code supplied by the borrower in their loan application\\
16 &            addr\_state &  The information submitted by the borrower in the loan application\\
17 &                   dti &  The ratio is determined by dividing the borrower's total monthly debt payments, excluding the mortgage and the requested LC loan, by the borrower's self-reported monthly income\\
18 &      earliest\_cr\_line &  The month when the borrower initially established their first reported credit line\\
19 &              open\_acc &  The month in which the borrower originally opened their initial documented credit account\\
20 &               pub\_rec &                Quantity of negative public records \\
21 &             revol\_bal &                     The overall balance of revolving credit \\
22 &            revol\_util &  The revolving line utilization rate refers to the proportion of credit that a borrower is currently utilizing in comparison to their total available revolving credit\\
23 &             total\_acc &  the overall count of credit lines presently recorded in the borrower's credit history\\
24 &   initial\_list\_status &  The loan's initial listing status \\
25 &      application\_type &  Specifies whether the loan application is for an individual or involves a joint application with two co-borrowers\\
26 &              mort\_acc &               Quantity of mortgage accounts        \\
27 &  pub\_rec\_bankruptcies &              Count of bankruptcies in public records\\
\bottomrule
\end{tabular}
\end{table}

\begin{figure}[H]
	\centering
	\includegraphics[width=\textwidth]{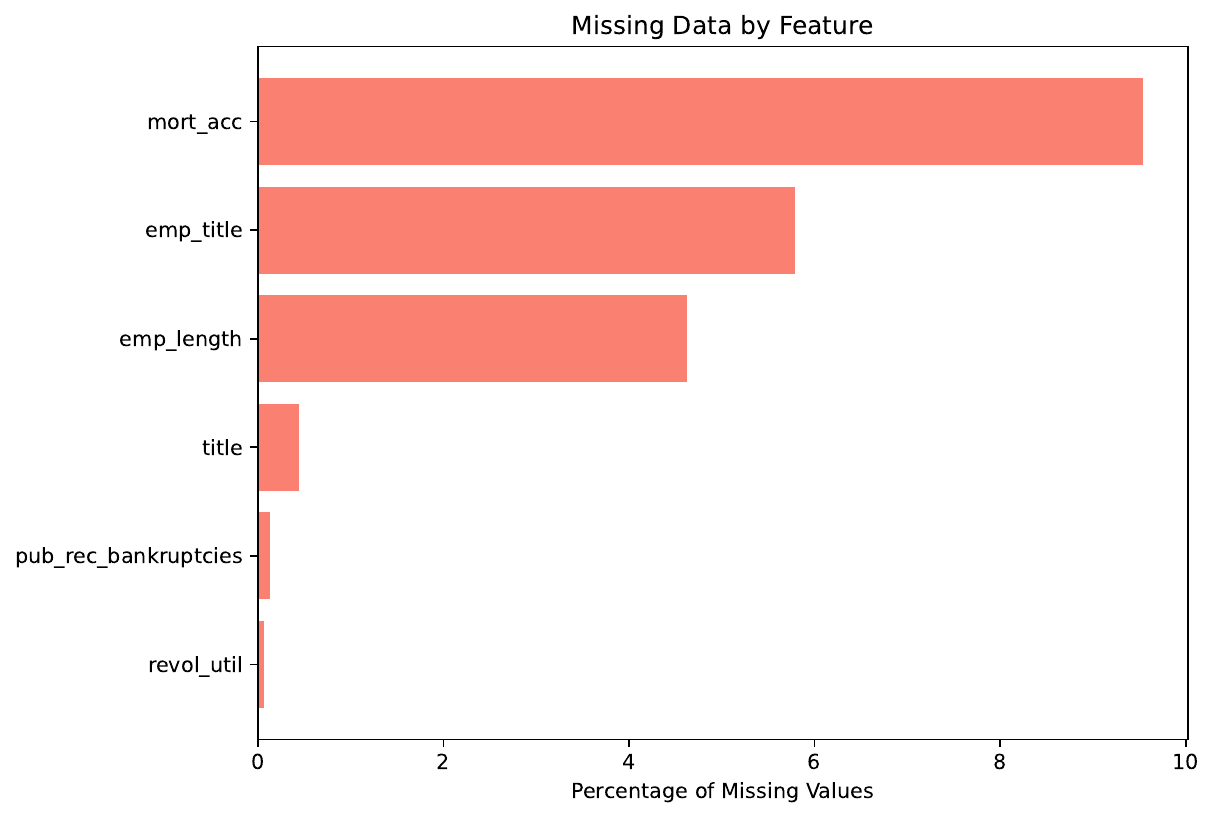}
	\caption{Distribution of missing values across features with gaps in the Lending Club dataset.}
 \label{fg:missing_data_plot}
\end{figure}

Table~\ref{table:var}  offers a brief description of each variable within the loan dataset.
The dataset consists of $396,030$ loan applications, each with $28$ features, which were disbursed through the LendingClub platform between June $2007$ and December $2016$.
The \enquote*{Missing data by feature} plot in Fig.~\ref{fg:missing_data_plot} visually illustrates the percentage of missing values across various features in the Lending Club dataset, with features sorted in ascending order by their missing percentages.
The feature with the least missing data is revol\_util, with only about $0.1\%$ missing, followed by pub\_rec\_bankruptcies at approximately $0.2\%$. 
The missing values increase significantly for title ($0.5\%$) and emp\_length ($6.0\%$), indicating more substantial gaps. 
Similarly, emp\_title has $4.5\%$ missing data, while mort\_acc shows the highest proportion of missing values at $9.0\%$, highlighting notable data gaps in these columns.

\noindent Although several variables contain missing observations, the overall proportion of missing values is relatively small, with the highest rate occurring in \texttt{mort\_acc} at approximately $9\%$. 
Given the large size of the dataset, observations containing missing values were removed rather than imputed. 
Importantly, this procedure involved removing incomplete records rather than discarding entire variables such as \texttt{mort\_acc} or \texttt{emp\_length}. 
After filtering, more than $80\%$ of the original observations remained available for analysis, which is sufficient to maintain the statistical representativeness of the dataset. 
Prior studies have shown that when missingness is limited and approximately random in large datasets, removing incomplete observations has a negligible impact on statistical inference \citep{bennett2001can,tabachnick2013using}.

From a modelling perspective, addressing missing data is a necessary preprocessing step before model training. 
Subsequent stages of the pipeline apply Recursive Feature Elimination (RFE) to identify the most informative predictors and Synthetic Minority Over-sampling Technique (SMOTE) to mitigate class imbalance, ensuring that the cleaned dataset provides reliable inputs for the proposed GWE framework.

\subsection{Numerical Feature Analysis}\label{sec:numF}

 The summary statistics presented in Table~\ref{table:summary} provide a detailed overview of the quantitative variables in the dataset, shedding light on their distributional properties and inherent characteristics.
 Each row captures key metrics, such as count, mean, standard deviation (std), and percentiles, collectively offering a robust statistical snapshot of the data.
The count column highlights the number of non-missing observations for each variable. 
The mean and std columns provide insights into central tendencies and variability.
For example, the mean loan\_amnt is approximately \$14,522, with a std of \$8,386, reflecting moderate variability around the mean. 
In contrast, the high standard deviation of variables such as annual\_inc  and revol\_bal indicates significant variability among borrowers, potentially influenced by a mix of low- and high-income earners or varying credit usage behaviors.
The percentile values (25\%, 50\%, and 75\%) further detail the distribution of each variable. 
For instance, pub\_rec\_bankruptcies demonstrates a heavily skewed distribution, with the first quartile (25\%), median (50\%), and third quartile (75\%) values all being zero.
This indicates that a majority of borrowers have no recorded public bankruptcies.
However, the maximum value of 8 highlights the existence of a small subset of borrowers with multiple bankruptcies, indicating significant variability within this minority group.
Outliers appear prominently in variables like annual\_inc and dti.
For annual\_inc, the mean income of \$75,817 starkly contrasts with the maximum value of \$8,706,582, suggesting the presence of extreme high-income outliers. 
Similarly, the dti variable shows an unusually high maximum of 380.53, far exceeding typical debt-to-income thresholds. These outliers could skew summary statistics and distort modelling outcomes if not properly addressed.
The minimum values for variables such as revol\_util  and dti are zero, indicating that some borrowers have no credit utilisation or reported debt. 
On the other hand, maximum values, such as $892.3$ for revol\_util and $380.53$ for dti, are uncharacteristically high, potentially signalling reporting errors, extreme behaviours, or specific subgroups within the dataset that warrant further investigation.
In terms of feature diversity, variables like total\_acc  and mort\_acc highlight variation in borrowers' credit profiles. 
For instance, while the median borrower has one mort\_acc, some have as many as $34$, underscoring differences in financial commitments.

\begin{table} [htbp]
\centering 
\caption{Summary statistics for loan default related variables.} 
\label{table:summary}
\begin{tabular}{l|rrrrrrrr}
\toprule
 & count & mean & std & min & 25\% & 50\% & 75\% & max \\
  \midrule
loan\_amnt & 335867.00 & 14522.76 & 8386.60 & 1000.00 & 8000.00 & 12400.00 & 20000.00 & 40000.00 \\
int\_rate & 335867.00 & 13.80 & 4.51 & 5.32 & 10.64 & 13.35 & 16.78 & 30.99 \\
installment & 335867.00 & 445.16 & 251.59 & 28.75 & 263.78 & 388.20 & 583.79 & 1533.81 \\
annual\_inc & 335867.00 & 75817.04 & 61972.24 & 5000.00 & 47000.00 & 65000.00 & 90000.00 & 8706582.00 \\
dti & 335867.00 & 17.72 & 8.15 & 0.00 & 11.63 & 17.25 & 23.42 & 380.53 \\
open\_acc & 335867.00 & 11.60 & 5.18 & 1.00 & 8.00 & 11.00 & 14.00 & 90.00 \\
pub\_rec & 335867.00 & 0.18 & 0.54 & 0.00 & 0.00 & 0.00 & 0.00 & 86.00 \\
revol\_bal & 335867.00 & 16193.01 & 20990.70 & 0.00 & 6302.00 & 11480.00 & 19990.00 & 1743266.00 \\
revol\_util & 335867.00 & 54.37 & 23.96 & 0.00 & 36.90 & 55.30 & 73.00 & 892.30 \\
total\_acc & 335867.00 & 25.86 & 11.90 & 2.00 & 17.00 & 24.00 & 33.00 & 151.00 \\
mort\_acc & 335867.00 & 1.80 & 2.14 & 0.00 & 0.00 & 1.00 & 3.00 & 34.00 \\
pub\_rec\_bankruptcies & 335867.00 & 0.13 & 0.36 & 0.00 & 0.00 & 0.00 & 0.00 & 8.00 \\
\bottomrule
\end{tabular}
\end{table}

\begin{figure}[H]
	\centering
	\includegraphics[width=\textwidth]{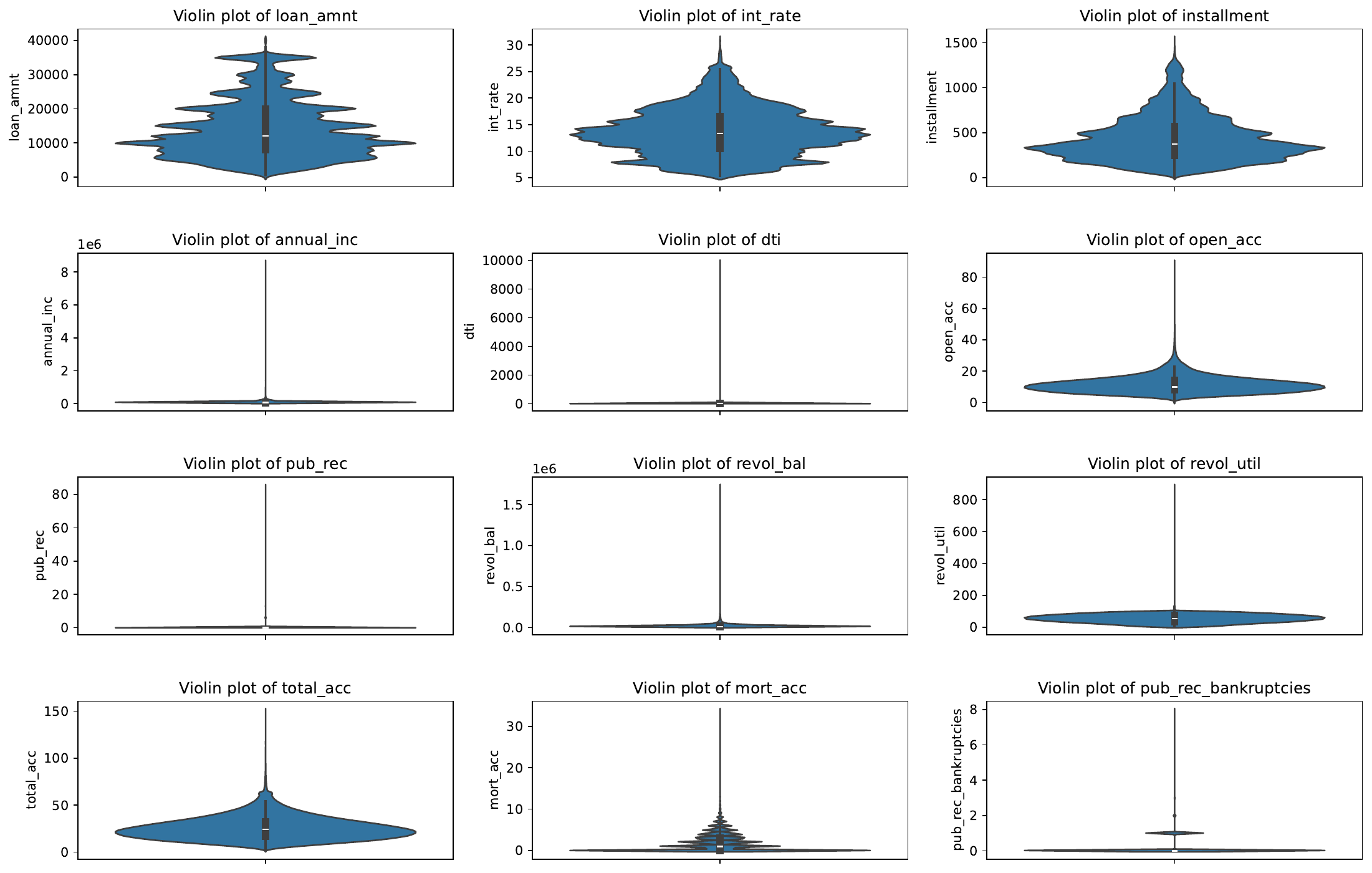}
	\caption{Violin plots showing the distribution and variability of key numerical features in the Lending Club dataset.
    The plots reveal the central tendency, spread, and potential outliers within each feature, providing insights into the characteristics of the borrowers and loans.}
 \label{fg:violin_plot_grid}
\end{figure}

The violin plots in Fig~\ref{fg:violin_plot_grid} visually depict the distribution of numerical variables in the Lending Club dataset, illustrating the central tendency, variability, and possible outliers. The distribution of variables like loan\_amnt, int\_rate, and installment is right-skewed, suggesting that most loans have lower amounts, interest rates, and installments, with some higher outliers. Similarly, annual\_inc is heavily right-skewed, with the majority of borrowers earning lower incomes, while a minor segment has considerably higher earnings. The dti plot also exhibits right skewness, reflecting that most borrowers possess lower debt-to-income ratios, despite a few extreme outliers. The open\_acc distribution is approximately symmetric, with a peak around 10--15 open accounts, whereas pub\_rec is heavily left-skewed, with most borrowers lacking public records, but a few having higher values. The revol\_bal distribution is right-skewed, indicating that most borrowers possess lower revolving balances, while a minority has significantly elevated balances. Revol\_util is more balanced, showing that most borrowers use a moderate amount of their available credit. The total\_acc distribution is right-skewed, with most borrowers having a moderate count of total accounts, and mort\_acc is similarly skewed, with most borrowers holding none or just one mortgage. Lastly, pub\_rec\_bankruptcies is left-skewed, with the majority having no bankruptcies, yet a few outliers denote those with a history of bankruptcies.
These observations highlight the presence of extreme values in several financial variables. 
Rather than removing these observations, which may represent legitimate high-income borrowers or unusual credit profiles, the modelling pipeline relies on robust ensemble learners and feature selection mechanisms to mitigate the influence of extreme values. 
This design choice preserves the diversity of borrower characteristics while reducing sensitivity to noise during model training.

\begin{figure}[H]
	\centering
	\includegraphics[width=\textwidth]{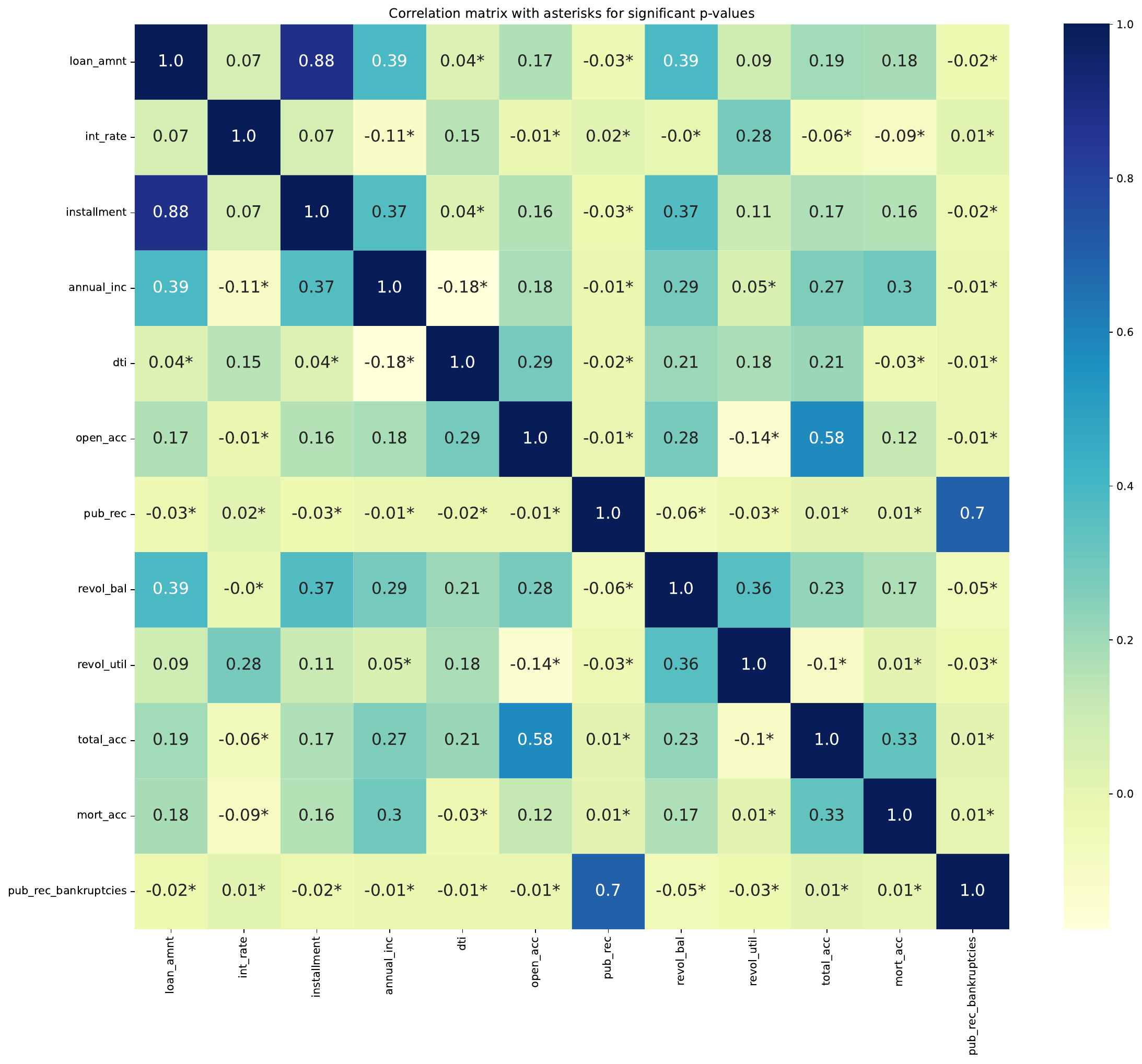}
	\caption{Heatmap of Pearson correlation coefficients among key numerical features in the Lending Club dataset, emphasising statistically significant relationships influencing loan default risk.}
 \label{fg:_cormatix_1}
\end{figure}

To explore the linear relationships between numerical features in the Lending Club dataset, we created a correlation matrix using Pearson's product-moment correlation coefficient formula:

\begin{equation} \label{eq:corr1}
\text{Corr}(U, V) = \frac{\text{cov}(U, V)}{\sigma_U \cdot \sigma_V},
\end{equation}  

\noindent where \(U\) and \(V\) represent any two numerical variables in the dataset, such as loan\_amnt, annual\_inc, int\_rate, and others. 
Here, \(\text{Corr}(U, V)\) measures the strength and direction of the linear relationship between \(U\) and \(V\), \(\text{cov}(U, V)\) is their covariance, and \(\sigma_U\) and \(\sigma_V\) are the stds of \(U\) and \(V\), respectively. 
By computing pairwise correlations, we created a matrix visualised in Fig~\ref{fg:_cormatix_1}, which highlights significant and weak associations among these features--helping to contextualize their roles in credit risk evaluation.
The presence of asterisks ($\ast$) denotes statistically significant correlations at conventional p-value thresholds, emphasising relationships that warrant deeper exploration.
A strong positive correlation ($0.88$) is observed between loan\_amnt and installment, which is intuitive since loan amounts directly determine installment sizes. This close relationship underscores the importance of considering these variables together in predictive models for loan repayment behaviour. 
Additionally, loan\_amnt shows moderate correlations with annual\_inc ($0.39$), revol\_bal ($0.39$), and total\_acc ($0.19$), indicating that higher loan amounts are often associated with higher incomes, larger revolving balances, and more extensive credit histories. 
These patterns suggest that wealthier borrowers or those with greater credit experience tend to access higher loan amounts, which may have implications for default risk.
The interest rate (int\_rate) shows a weak but significant correlation with revol\_util ($0.28$) and dti ($0.15$), indicating that borrowers with higher utilization rates or debt-to-income ratios may face higher interest rates, potentially reflecting lenders' risk-adjusted pricing strategies. 
However, its negative correlation with annual\_inc ($-0.11$) suggests that higher-income borrowers are offered lower interest rates, likely due to their perceived lower risk. 
These dynamics highlight how creditworthiness factors interact with interest rates, influencing the cost of borrowing and potentially affecting repayment behaviour.
Debt-to-income ratio  is positively correlated with open\_acc ($0.29$), revol\_bal ($0.21$), and total\_acc ($0.21$). 
These relationships indicate that borrowers with higher credit activity or balances tend to have higher debt burdens, which could elevate their default risk.
Conversely, dti exhibits a weak but negative correlation with annual\_inc ($-0.18$), suggesting that higher-income borrowers generally maintain lower debt ratios, reflecting better financial stability.
Public records (pub\_rec) and bankruptcies (pub\_rec\_bankruptcies) exhibit a strong correlation ($0.7$), as expected, given their overlap in representing adverse credit events.
Both variables are negatively correlated with most other financial metrics, indicating that borrowers with negative credit histories typically have less access to credit.
For instance, pub\_rec is weakly but negatively correlated with loan\_amnt ($-0.03$) and installment ($-0.03$), suggesting that borrowers with public records are less likely to secure substantial loans.
The variable revol\_bal shows moderate positive correlations with loan\_amnt ($0.39$), installment ($0.37$), and annual\_inc ($0.29$), indicating that borrowers with higher revolving balances tend to take on larger loans and often have higher incomes.
This alignment suggests that revolving balances may serve as a proxy for borrowing capacity and financial behaviour. 
Similarly, revol\_util is positively correlated with int\_rate ($0.28$), highlighting that higher credit utilisation rates are associated with higher borrowing costs, which could influence default risks.
The analysis also reveals that mort\_acc is positively correlated with total\_acc ($0.33$) and annual\_inc ($0.3$), suggesting that borrowers with higher incomes or more extensive credit histories are more likely to have mortgage accounts.
This reflects the interconnected nature of credit variables and the potential role of mortgages in shaping overall financial behaviour.
The strong correlation observed between certain variables, particularly \texttt{loan\_amnt} and \texttt{installment}, suggests potential redundancy among predictors. 
Instead of manually removing correlated features, this study employs Recursive Feature Elimination (RFE) in the subsequent modelling stage. 
RFE systematically evaluates the predictive contribution of each variable and removes redundant predictors based on model performance, allowing the final feature set to be determined through data-driven selection rather than purely statistical correlation.

\subsection{Categorical Feature Analysis}\label{sec:catF}

Before model training, categorical variables such as \texttt{home\_ownership}, 
\texttt{purpose}, \texttt{verification\_status}, and \texttt{application\_type} 
were converted into numerical representations to ensure compatibility with 
machine learning algorithms that require numerical inputs. 
Specifically, One-Hot Encoding was applied during preprocessing to transform 
categorical variables into binary indicator features while avoiding the introduction of ordinal relationships between categories. 
This transformation enables models such as Logistic Regression and Support 
Vector Machines to process categorical information appropriately, while also 
remaining compatible with tree-based ensemble models used in this study.

\begin{figure}[H]
	\centering
	\includegraphics[width=\textwidth]{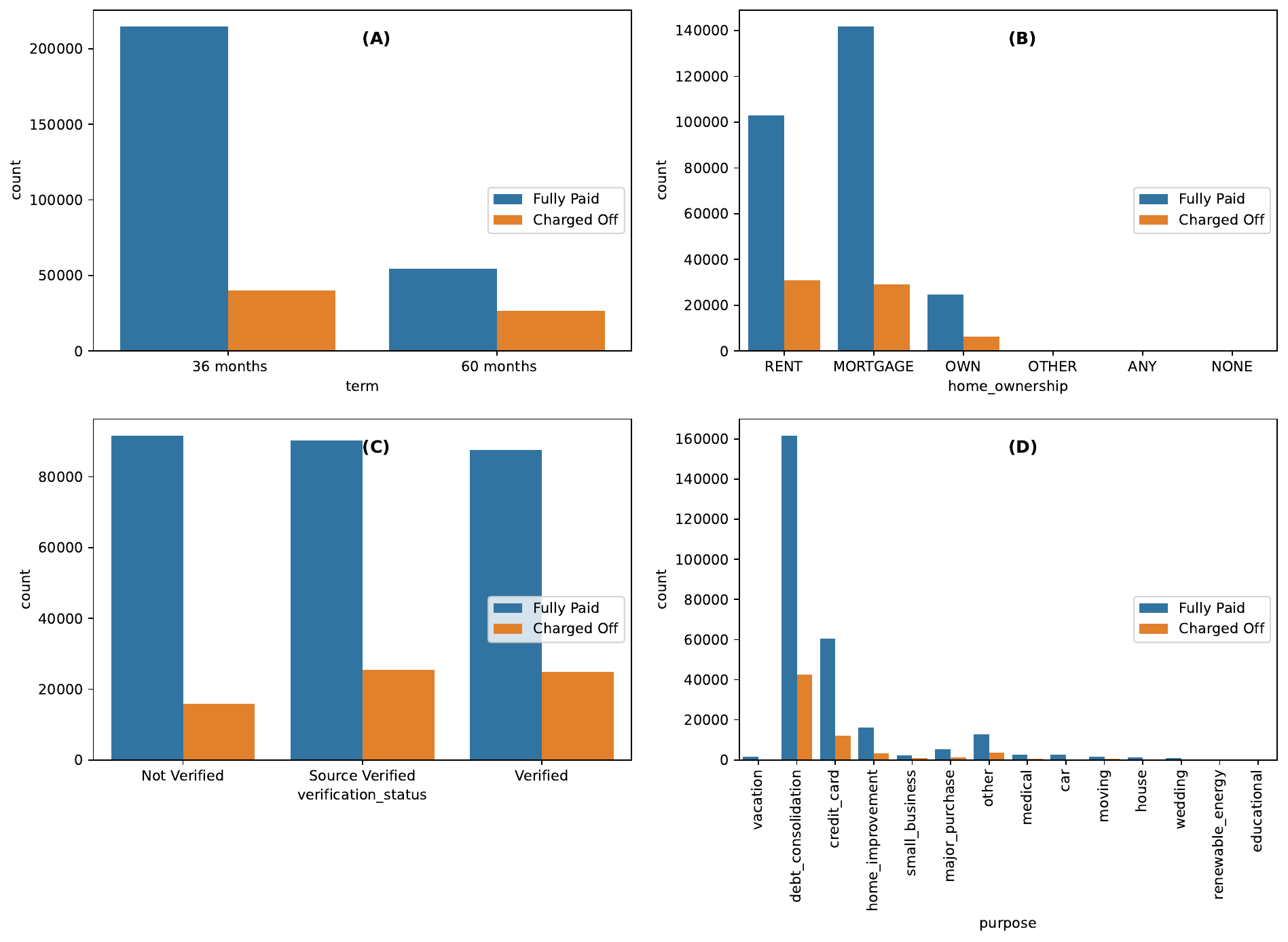}
	\caption{Comparison of loan outcomes--fully paid or charged off-based on borrower-provided  number of payments (Panel A), home ownership status (Panel B), loan verification status (Panel C), and loan request  category (Panel D).}
 \label{fg:compare_ls}
\end{figure}

Referring to Fig.~\ref{fg:compare_ls} A, the count of payments on a loan with a shorter duration, for example, 36 months, might surpass those of a longer-term loan, such as 60 months.
Typically, shorter-term loans necessitate larger monthly payments, enabling borrowers to settle both the principal and interest in a shorter span.
Consequently, these loans often demand larger individual payments compared to longer-term options, which may feature smaller monthly dues. Moreover, borrowers dealing with shorter-term loans generally incur lower total interest over the loan's life.
A notable chunk of each payment contributes to diminishing the principal balance, hastening the loan's repayment. 
This reduction in interest expenses serves as an appealing factor for borrowers.
Fig.~\ref{fg:compare_ls} B shows how various statuses of home ownership, such as mortgage, rent, own, and others, affect the financial traits, responsibilities, and motivations of the borrowers regarding the repayment of the full loan versus experiencing loan charges.
Those with a mortgage, symbolising property ownership, often exhibit heightened financial stability and dedication. 
Having made a significant, long-term investment in a home reflects responsible financial behaviour.
On the contrary, renters and individuals with unconventional home ownership may showcase reduced financial stability and commitment, possibly due to greater mobility and lesser attachment to their current residence. This might make them less inclined to prioritise loan payments.
Although verification status holds importance, loans being charged off might be attributed to several reasons unrelated to the verification process, such as financial instability, changing circumstances, or unforeseen events. 
Therefore, the verification status might have less impact in such cases, as evidenced in Fig.~\ref{fg:compare_ls} C. 
The observation regarding the \enquote{debt consolidation} category, as depicted in Fig.~\ref{fg:compare_ls} D, showcasing a higher count of both fully paid and charged-off loans, can be linked to multiple factors related to borrower behaviour and financial situations.
For example, borrowers seeking debt consolidation loans may carry larger outstanding debts compared to those seeking loans for specific purposes like vacations or medical expenses. This scenario could result in both higher counts of fully paid and charged-off loans within the category.

\begin{figure}[H]
	\centering
	\includegraphics[width=\textwidth]{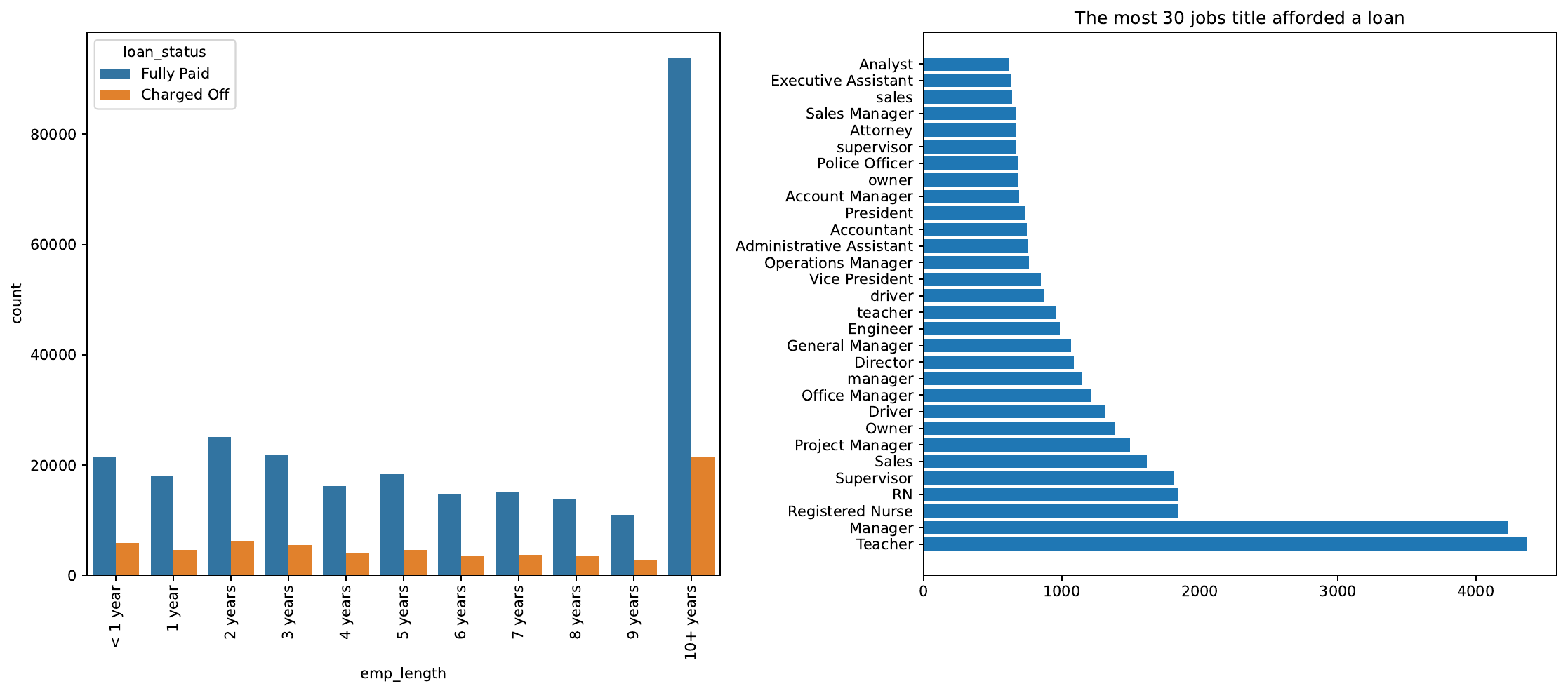}
	\caption{The left panel shows the distribution of fully paid and charged-off loans across different employment lengths. The right panel displays the top $30$ job titles that were afforded a loan, highlighting the variety of professions represented in the dataset.}
 \label{fg:emp_title_ls}
\end{figure}

The left graph in Fig.~\ref{fg:emp_title_ls} illustrates the connection between the duration of employment and the rates of loan defaults. It clearly shows that borrowers with a more extended employment tenure often exhibit reduced default rates. This pattern aligns with the notion that a longer employment record typically indicates greater financial stability and a lower risk profile. 
The right graph showcases the distribution of loan approvals across various job titles. Interestingly, certain job titles such as Teacher, Registered Nurse, and Manager exhibit higher loan approval frequencies. This indicates that these professions are associated with lower observed default rates, likely due to stable employment and income.

\section{Pre-processing Techniques} \label{sec:prep}

This section describes the preprocessing pipeline applied to the Lending Club dataset prior to model training. 
The objective of this stage is to transform the raw loan records into a clean and informative feature set suitable for machine learning models. 
The preprocessing workflow consists of three main stages: detection and correction of extreme observations, mitigation of class imbalance through statistical augmentation, and identification of the most informative predictors using RFE. 
These steps form the data preparation pipeline that generates the final model inputs used by the proposed GWE framework.

\subsection{A Hybrid Outlier Detection and Correction Pipeline}\label{subsec:BCP-HI}

The financial lending data often exhibits structural changes, such as abrupt shifts in trends due to macroeconomic events or policy changes.
Detecting these structural changes is critical, as they signify segments where anomalous behavior might occur.
Bayesian Change Point (BCP) Analysis, as introduced by \citet{barry1993bayesian},  serves as the first step in the pipeline.
This method identifies change points by evaluating the likelihood of observing data segments under different distributional assumptions, such as normal or uniform. 
For instance, BCP can be used to identify changes in interest rates over time, helping to understand shifts in monetary policy and their impact on lending behaviour.
Mathematically, for a segment \( \mathbf{y} = \{y_1, y_2, \dots, y_n\} \), the BCP analysis computes the combined likelihood \( \mathcal{L} \) of splitting the data into two subsets, \( \mathbf{y_1} \) and \( \mathbf{y_2} \). The log-likelihoods are given as:
\begin{equation}\label{eq:bcp}
\mathcal{L} = \sum_{i=1}^{n_1} \log P(y_i | \mu_1, \sigma_1) + \sum_{j=1}^{n_2} \log P(y_j | \mu_2, \sigma_2),
\end{equation}
where \( P(y | \mu, \sigma) \) represents the probability density function of the assumed distribution.
It should be noted that the Lending Club dataset is cross-sectional rather than a temporal time series. 
In this study, BCP analysis is therefore used as a statistical segmentation tool applied to ordered observations to identify abrupt distributional shifts or irregular patterns in numerical variables. 
The objective is not to model temporal dynamics but to detect segments where anomalous behaviour may occur.
A prior probability is incorporated to adjust for the likelihood of specific segmentations, forming a pseudo-posterior that identifies the most likely change points.
This analysis divides the dataset into coherent segments, ensuring that the subsequent outlier detection methods are applied contextually to homogeneous regions of the data. This is particularly relevant in financial datasets where localised patterns may differ across segments.

After identifying change points, the Interquartile Range (IQR) technique is applied to detect global outliers within each segment. Outliers are defined as data points lying beyond a threshold determined by the IQR:
\begin{equation}\label{eq:iqr_0}
\text{IQR} = Q_3 - Q_1,
\end{equation}
where \( Q_1 \) and \( Q_3 \) are the first and third quartiles, respectively. Data points outside the range:
\begin{equation}\label{eq:iqr_1}
[Q_1 - k \times \text{IQR}, Q_3 + k \times \text{IQR}]
\end{equation}
are flagged as outliers \citep{tukey1977eda}. 
The constant \( k \) is typically set to 1.5 but can be adjusted (e.g., \( k = 3 \)) to balance sensitivity and specificity, as done in this study. 
This approach provides a foundational layer for identifying extreme values that deviate significantly from the central tendency of the data, complementing the more localised analysis provided by the Hampel Identifier \citep{aggarwal2017introduction}.

To refine outlier detection within the segments identified by BCP, the Hampel Identifier is employed.
This method leverages the Median Absolute Deviation (MAD) to robustly identify outliers in non-Gaussian distributions \citep{hampel1974robust}.
For a data point \( y_i \), the Hampel Identifier calculates:
\begin{equation}\label{eq:mad_0}
\text{MAD} = \text{median}(|y_j - \text{median}(y)|), \quad j = 1, 2, \dots, n,
\end{equation}
and flags \( y_i \) as an outlier if:
\begin{equation}\label{eq:mad_1}
|y_i - \text{median}(y)| > k \times \text{MAD}.
\end{equation}
The threshold \( k \) is chosen based on the desired level of sensitivity. By focusing on individual segments, the Hampel Identifier accounts for localised patterns and ensures that outliers are detected within the appropriate context.
Recent studies, such as those by \citet{ansah2024enhancing}, validate the Hampel Identifier's effectiveness in improving data quality for machine learning models, emphasising its role in preprocessing stages.

Financial datasets, such as those from the Lending Club, often exhibit high-dimensional feature spaces due to the large number of loan attributes and numerous instances. To enhance interpretability, reduce computational complexity, and address issues like multicollinearity, dimensionality reduction techniques such as Principal Component Analysis (PCA) are integrated into the pipeline.
For instance, \citet{jolliffe2002principal} demonstrated PCA's utility in extracting meaningful structures from complex datasets, making it a cornerstone of this methodology. 
In this study, PCA is used only as an exploratory dimensionality reduction technique to examine the structure of the high-dimensional feature space. 
The outlier detection procedures described above (BCP, IQR, and the Hampel identifier) are applied in the original feature space rather than in the reduced PCA space. 
PCA is therefore used solely to visualise dominant variance patterns and to inspect potential correlations or clustering structures among variables. 

Mathematically, PCA transforms the original feature matrix $\mathbf{X}$ into a lower-dimensional representation by projecting the data onto orthogonal directions of maximum variance. 
This is achieved by computing the covariance matrix $\Sigma$ of the standardised data and performing eigenvalue decomposition to obtain eigenvectors $\mathbf{W}$ corresponding to the largest eigenvalues. 
The reduced representation is then obtained as
\begin{equation}\label{eq:pca}
\mathbf{Y}_{\text{reduced}} = \mathbf{X}\mathbf{W}_k ,
\end{equation}
where $\mathbf{W}_k$ contains the $k$ eigenvectors associated with the largest eigenvalues.
By reducing the dimensionality of the lagged data, PCA enables the detection pipeline to focus on the most salient temporal patterns while suppressing noise and redundant information. This is particularly beneficial for anomaly detection, as it enhances the separation of anomalous patterns from regular ones in the reduced space.
The choice of \( k \), the number of retained components, is determined by the cumulative explained variance. For example, retaining components that explain $95\%$ of the variance ensures a balance between preserving data fidelity and maintaining computational efficiency.
After detecting outliers, the final step involves correcting them to mitigate their impact on predictive models.
A median filter, a robust smoothing technique, is applied.
For each outlier, a sliding window of size \( w \) centred on the outlier is used to compute the median of neighbouring values:
\begin{equation}\label{eq:corr}
y_i^{\text{corrected}} = \operatorname{median}\big( y_{i-\lfloor w/2\rfloor},\dots,y_i,\dots,y_{i+\lfloor w/2\rfloor}\big),.
\end{equation}
This approach preserves the overall data structure while addressing local anomalies, ensuring that corrected values remain representative of their immediate context.
The pseudocode in Algorithm~\ref{al:bcp-hybrid} highlights the systematic integration of Bayesian methods and statistical techniques to detect and correct outliers in financial datasets exhibiting structural changes.

\begin{figure}[H]
	\centering
	\includegraphics[width=\textwidth]{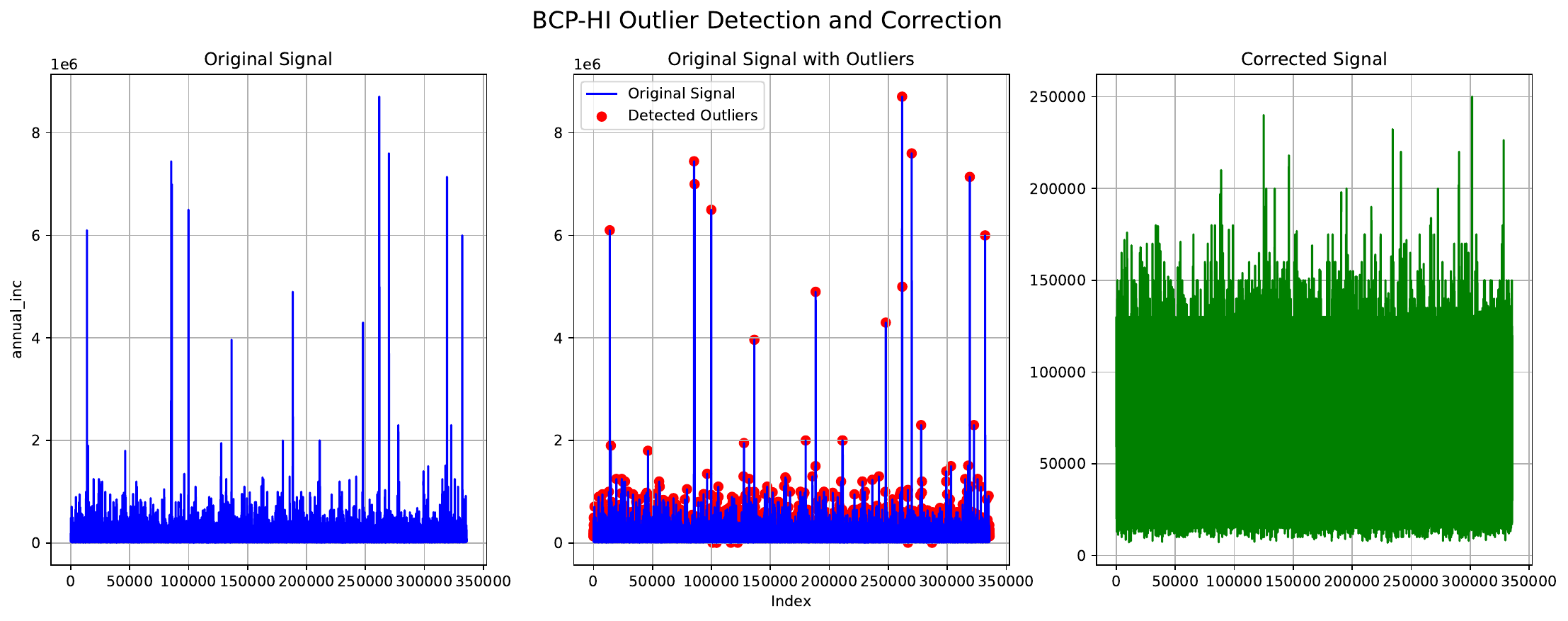}
	\caption{Outlier Detection and Correction using the BCP-HI Pipeline: (Left) The original signal (annual\_inc) showing significant spikes indicative of potential outliers. (Middle) The original signal with detected outliers marked as red dots. (Right) The corrected signal after outlier detection and correction displays a smoother, more reliable trend for further analysis.}
 \label{fg:BCP-HI}
\end{figure}

\begin{algorithm}[H]
\caption{A Bayesian and Statistical Hybrid Pipeline for Outlier Detection and Correction}\label{al:bcp-hybrid}
\begin{algorithmic}[1]
\Require Dataset $\mathbf{y} = \{y_1, y_2, \dots, y_N\}$
\Ensure Corrected dataset $\mathbf{y}^{\text{corrected}}$
\State \textbf{Initialize} segment set $\mathcal{S} = \{\}$ and corrected dataset $\mathbf{y}^{\text{corrected}} = \mathbf{y}$
\State \textbf{Step 1: Change Point Detection using Bayesian Change Point Analysis}
\State Compute likelihood $\mathcal{L}$ for each potential segmentation of $\mathbf{y}$ (Equation \ref{eq:bcp})
\State Incorporate priors to identify the most probable change points
\State Divide $\mathbf{y}$ into segments $\mathcal{S} = \{\mathbf{y}_1, \mathbf{y}_2, \dots, \mathbf{y}_k\}$

\State \textbf{Step 2: Global Outlier Detection using IQR}
\For{each segment $\mathbf{y}_i \in \mathcal{S}$}
    \State Compute $Q_1$, $Q_3$, and $\text{IQR} = Q_3 - Q_1$ (Equation \ref{eq:iqr_0})
    \State Identify outliers: $y \notin [Q_1 - k \times \text{IQR}, Q_3 + k \times \text{IQR}]$ (Equation \ref{eq:iqr_1})
    \State Flag global outliers within $\mathbf{y}_i$
\EndFor

\State \textbf{Step 3: Localised Outlier Detection using Hampel Identifier}
\For{each segment $\mathbf{y}_i \in \mathcal{S}$}
    \State Compute $\text{median}(\mathbf{y}_i)$ and $\text{MAD}$ (Equation \ref{eq:mad_0})
    \State Identify localized outliers: $|y_j - \text{median}(\mathbf{y}_i)| > k \times \text{MAD}$ (Equation \ref{eq:mad_1})
    \State Flag localized outliers within $\mathbf{y}_i$
\EndFor

\State \textbf{Step 4: Dimensionality Reduction with PCA}
\State Perform PCA on $\mathbf{y}$: $\mathbf{Y}_{\text{reduced}} = \mathbf{X} \mathbf{W}$ (Equation \ref{eq:pca})
\State Use reduced features to identify clusters and visualise anomalies

\State \textbf{Step 5: Outlier Correction using Median Filter}
\For{each flagged outlier $y_i$ in $\mathbf{y}$}
    \State Compute sliding window median: $y_i^{\text{corrected}} = \text{median}(y_{i-w/2:i+w/2})$ (Equation \ref{eq:corr})
    \State Update $\mathbf{y}^{\text{corrected}}$
\EndFor

\State \Return Corrected dataset $\mathbf{y}^{\text{corrected}}$
\end{algorithmic}
\end{algorithm}

\noindent The results of the Bayesian Changepoint Hybrid Integration (BCP-HI) Outlier Detection and Correction Pipeline are displayed in Fig.~\ref{fg:BCP-HI}, which consists of three panels representing different stages of the data processing.
In the first panel, the Original Signal shows the raw financial data (annual\_inc) plotted against the index. 
The signal reveals significant spikes that deviate drastically from the general trend, which indicates the presence of potential outliers. These anomalies may be due to data errors, extreme events, or sudden structural changes, which could distort subsequent analyses and lead to misleading conclusions.
The second panel shows the original signal with detected outliers marked as red dots.
The outlier detection step, implemented by the BCP-HI pipeline, accurately identifies these extreme points, which correspond to the sharp spikes in the first panel. These outliers are crucial because they could skew financial models, particularly in datasets where data accuracy is paramount. Detecting and addressing these anomalies is important, especially in the context of financial datasets, where outliers can represent rare but significant events such as market shocks or operational issues that may not follow the usual patterns.
Finally, the Corrected Signal in the third panel presents the data after the outlier correction step. The significant spikes have been removed, resulting in a smoother, more consistent trend. This corrected signal is now much more stable and reliable for use in further modelling tasks. The outlier correction ensures that the dataset reflects more accurate, normal financial conditions without the influence of erratic data points, making it suitable for predictive analysis or decision-making processes. By applying this correction, the dataset becomes more robust, reducing the risk of overfitting or misinterpretation in the modelling.
This stage of the pipeline is essential in the preprocessing of the Lending Club data because it directly addresses the problem of outliers, which can heavily influence the performance and accuracy of predictive models. 
In financial datasets like those from Lending Club, outliers may be caused by rare but impactful events, such as sudden market shifts or data entry mistakes. These outliers, if not addressed, can distort models that rely on more predictable patterns. For research focusing on financial outcomes or predicting loan defaults, ensuring that the data used for training models are free of extreme and misleading values is crucial to develop reliable and interpretable results. This preprocessing stage supports the research's broader goal of building predictive models that can accurately forecast financial trends and loan performance. By addressing outliers in the dataset, the pipeline ensures that the data used for further analysis is robust and free from distortions caused by extreme values. This step is critical in ensuring that the predictive models built on this data are both reliable and interpretable.
We apply the same approach to the other numerical features in the Lending Club dataset, as shown in Fig~\ref{fg:violin_plot_grid}, to ensure consistency and accuracy across all relevant financial variables.

\subsection{Data Augmentation Technique for Class Imbalance}\label{subsec:dimbearn}

\begin{figure}[H]
\begin{subfigure}[b]{0.5\textwidth}
    \centering
    \includegraphics[width=\textwidth]{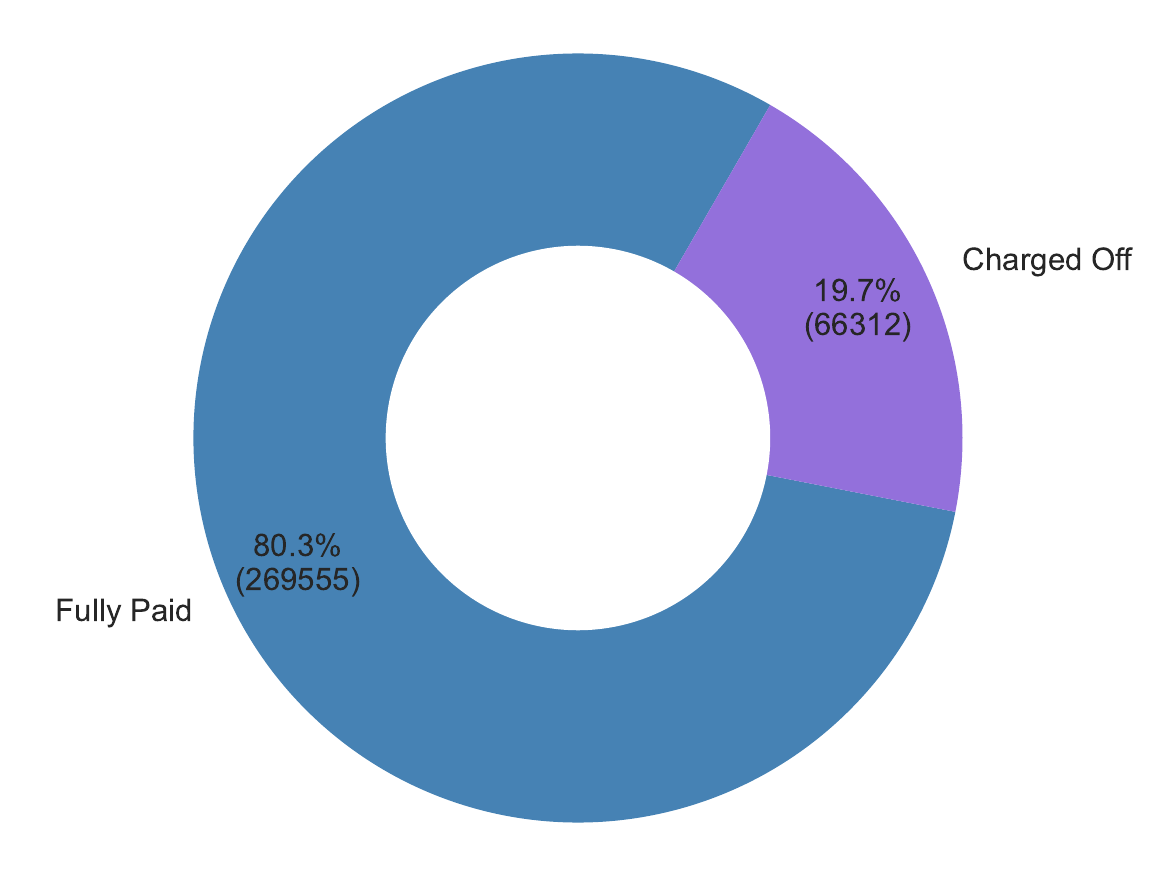}
    \caption{}
    \label{figs:baised}
\end{subfigure}%
\quad
\begin{subfigure}[b]{0.5\textwidth}
    \centering
    \includegraphics[width=\textwidth]{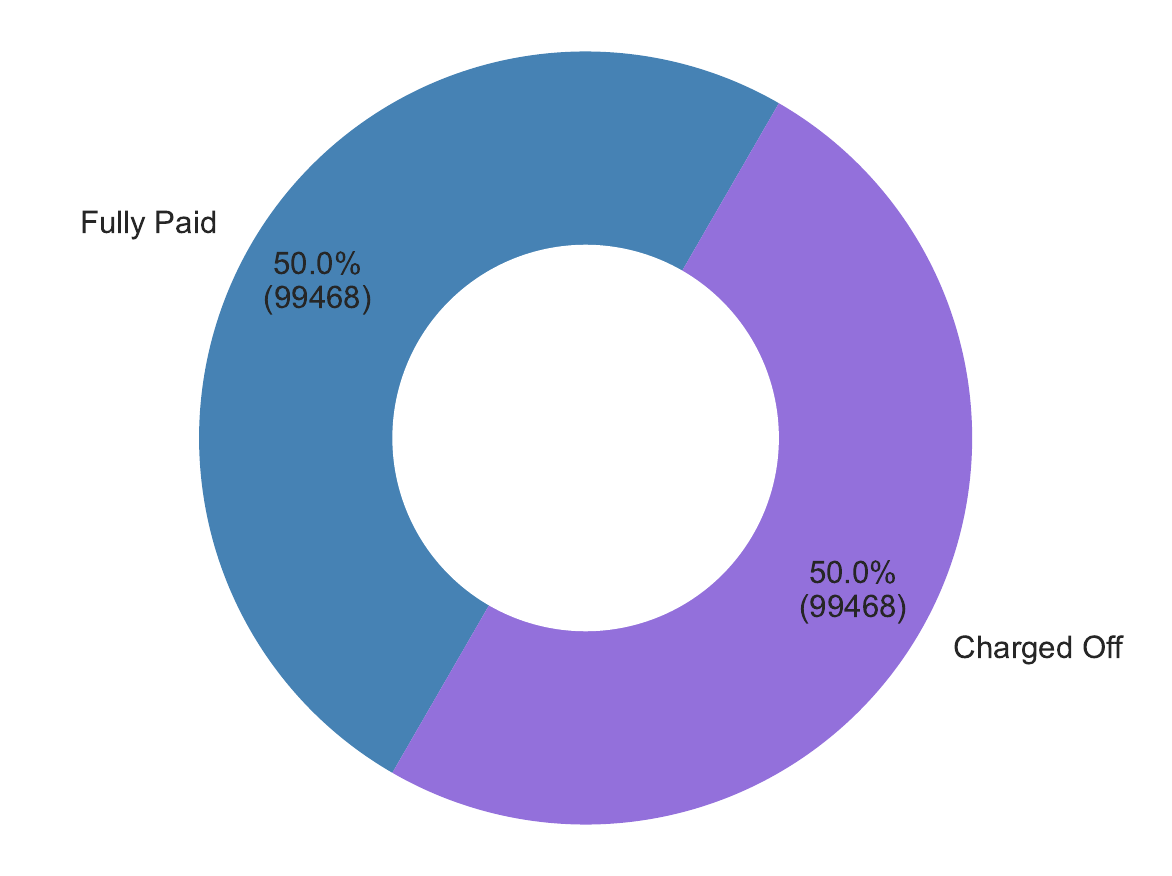}
    \caption{ }
    \label{figs:unbaised}
\end{subfigure}        
\caption{Comparison of loan status plots: (a) illustrates the biased plot, while (b) showcases the unbiased plot, highlighting the impact of addressing data imbalance in depicting accurate loan status distributions.}
\label{fig:resample}
\end{figure}

The loan status in the Lending Club dataset is highly imbalanced, as illustrated in the left panel of Fig.~\ref{figs:baised}, where the Charged Off instances amount to 66,312 ($\sim 19.7\%$), while the Fully Paid cases dominate at 269,555 ($\sim 80.3\%$). 
This severe class imbalance is typical in financial datasets, where the minority class, often representing defaults or adverse events, is of primary interest \citep{burez2009handling,he2009learning}.
Predictive models trained on such unbalanced datasets tend to perform poorly on the minority class, as they are biased toward the majority class due to the skewed data distribution. This imbalance poses significant challenges in financial decision-making, where accurately identifying high-risk loans is critical \citep{sun2007cost}.

To address these challenges, there is a pressing need for robust data augmentation techniques that enhance the representation of the minority class without distorting the dataset's statistical properties. In this work, the adopted data augmentation approach employs quantile transformation, controlled undersampling, and Gaussian noise-based augmentation to address class imbalance, ensuring the minority class is effectively represented in the dataset. This method is grounded in statistical and mathematical principles tailored to preserve the original data's integrity while introducing sufficient variability to improve model robustness. When compared to widely used techniques such as the Synthetic Minority Oversampling Technique (SMOTE) and centroid-based methods, this approach demonstrates greater flexibility, interpretability, and control over the augmentation process. 

Quantile transformation is a critical preprocessing step that ensures numerical features follow a Gaussian distribution, addressing the issue of skewed or non-normal distributions common in real-world datasets.
This transformation utilises the empirical cumulative distribution function (ECDF) of each feature and maps it to a target normal distribution through an inverse cumulative distribution function (CDF). 
Mathematically, for a feature \(x\), the transformed value \(z\) is given by 
$
z = \Phi^{-1}[F(x)],
$
where \(F(x)\) is the ECDF of \(x\), and \(\Phi^{-1}\) is the inverse CDF of the normal distribution. Unlike standard scaling or min-max normalization, which are sensitive to outliers and assume linear relationships between features, quantile transformation preserves the rank and non-linear relationships among features, making it particularly robust for datasets with extreme values or complex distributions \citep{garcia2015data}.
This standardisation enhances the stability of subsequent steps, such as noise-based augmentation, and ensures that the added noise integrates seamlessly into the transformed distribution.

Undersampling the majority class is employed to mitigate the dominance of the majority class while maintaining an informative dataset.
By selecting a subset of majority samples at a controlled ratio ($1.5:1$ relative to the minority class), the technique avoids the pitfalls of information loss associated with more aggressive undersampling methods. 
Formally, the majority class subset \(D_{majority}^{sampled}\) is chosen such that \(|D_{majority}^{sampled}| = 1.5 \times |D_{minority}|\). This controlled balance ensures that the dataset remains representative while reducing the computational overhead of training models on highly imbalanced data.
The stratified undersampling approach preserves the diversity of the majority class and prevents the removal of critical patterns that might be essential for accurate classification.

To augment the minority class, Gaussian noise is introduced to existing samples, generating synthetic data points that retain the underlying structure and variability of the original data. 
For a given feature \(x\), the augmented values \(x'\) are computed as \(x' = x + \epsilon\), where \(\epsilon\) is random noise drawn from a normal distribution \(\mathcal{N}(0, \sigma_x^2)\) and \(\sigma_x = 0.05 \times \text{std}(x)\). The scaling factor of $0.05$ ensures that the perturbations are meaningful yet subtle enough to avoid distorting the data's statistical properties.
By directly adding noise rather than extrapolating between data points, as in SMOTE, this method avoids creating unrealistic samples, particularly in high-dimensional or noisy datasets.
It introduces variability that enhances model generalisation while preserving the original class semantics \citep{shorten2019survey}.

Compared to other augmentation techniques, this approach offers several advantages. 
SMOTE, for example, interpolates between nearest neighbors to generate synthetic samples.
While effective in some cases, it can create unrealistic data in complex or noisy feature spaces and distort feature distributions \citep{chawla2002smote}. 
Similarly, centroid- or cluster-based augmentation methods assume well-separated clusters or representative centroids, assumptions that may not hold in datasets with overlapping or irregular distributions. 
The Gaussian noise-based augmentation avoids these limitations by directly introducing controlled randomness aligned with the feature's intrinsic variability.
Quantile transformation further enhances this process by ensuring that the added noise conforms to a Gaussian distribution, preserving statistical fidelity.
The proposed approach also mitigates overfitting risks commonly associated with traditional oversampling methods, such as random oversampling, which merely duplicates minority samples.
By introducing diversity through controlled noise, the augmented dataset remains robust while avoiding redundancy. Furthermore, the simplicity of Gaussian noise addition ensures interpretability, a crucial factor in domains such as finance, where understanding the data generation process is as important as the predictive model itself.
In contrast, the interpolation and extrapolation mechanisms of SMOTE may obscure the origins of synthetic samples, making them less interpretable.

The pseudocode in Algorithm~\ref{al:augmentation} outlines the process adopted to enhance the minority class, mitigating the bias present in the Lending Club dataset. As shown in Fig.~\ref{figs:unbaised}, the augmentation process results in an equal representation of fully paid and charged-off loans, addressing the inherent class imbalance effectively.

\begin{algorithm}
\caption{Minority Class Enhancement Algorithm Using Statistical Data Augmentation}\label{al:augmentation}
\begin{algorithmic}[1]
\Require Dataset \( D = \{(X_i, Y_i)\}_{i=1}^N \), where \( X_i \) are features and \( Y_i \) is the target class label
\Ensure Augmented dataset \( D_{aug} \)

\State \textbf{Initialize} \( D_{minority} = \{(X_i, Y_i) \mid Y_i = \text{minority class}\} \)
\State \textbf{Initialize} \( D_{majority} = \{(X_i, Y_i) \mid Y_i = \text{majority class}\} \)
\State Set target ratio \( r = 1.5 \)
\State \textbf{Initialize} \( D_{aug} = D \)

\Comment{Step 1: Quantile Transformation}
\State Apply quantile transformation to numerical features in \( D \) to map them to a normal distribution:
\[
X' = \Phi^{-1}(F(X)),
\]
where \( F(X) \) is the empirical cumulative distribution function (ECDF) and \( \Phi^{-1} \) is the inverse CDF of the normal distribution.

\Comment{Step 2: Undersampling the Majority Class}
\State Compute \( n_{minority} = |D_{minority}| \) and \( n_{majority} = r \cdot n_{minority} \)
\State Randomly sample \( D_{majority}^{sampled} \subseteq D_{majority} \) such that \( |D_{majority}^{sampled}| = n_{majority} \)

\Comment{Step 3: Gaussian Noise-Based Augmentation}
\State Compute standard deviation \( \sigma_x \) for each numerical feature \( X \) in \( D_{minority} \)
\State Define noise \( \epsilon \sim \mathcal{N}(0, 0.05 \cdot \sigma_x^2) \)
\For{each sample \( (X_i, Y_i) \in D_{minority} \)}
    \State Generate augmented samples \( X_i' = X_i + \epsilon \)
    \State Add \( (X_i', Y_i) \) to \( D_{aug} \)
\EndFor

\Comment{Combine Augmented Dataset}
\State \( D_{aug} \gets D_{aug} \cup D_{majority}^{sampled} \)
\end{algorithmic}
\end{algorithm}

\subsection{Feature Selection}\label{subsec:fselect}

To comprehend the key elements influencing a borrower's probability of default, feature analysis plays a crucial role in forecasting loan defaults. 
This analytical approach helps financial firms uncover critical loan default indicators necessary for efficient charged-off assessment in lending. 
Studies like \citet{altman1968financial} on financial ratios and bankruptcy prediction serve as examples of this process in action. 
Moreover, \citet{chen2006} have emphasised the significance of the feature importance in predicting bank financial failures. 
This transparency is critical for regulatory compliance and building stakeholder confidence.

In this investigation, we aimed to enhance the model's effectiveness by disregarding less informative features and concentrating on those exhibiting substantial predictive capabilities. It is important to note that irrelevant input data has the potential to perplex algorithms, leading them to draw inaccurate inferences and consequently yield suboptimal outcomes.
Hence, we employed the Recursive Feature Elimination (RFE) method in conjunction with five ensemble algorithms based on decision trees: Extremely Randomized Trees (ERT) \citep{geurts2006extremely}, Light Gradient Boosting Machine (LightGBM) \citep{ke2017lightgbm}, Extreme Gradient Boosting (XGBoost) \citep{chen2016xgboost}, Random Forest (RF) \citep{breiman2001random}, and CatBoost \citep{prokhorenkova2018catboost}. This approach allowed us to identify the ten most influential features from a pool of $26$. 
Subsequently, these selected features were employed for further analysis in machine learning applications.

The RFE technique is employed to enhance the efficiency and interpretability of ensemble models.
It operates by iteratively fitting the model and eliminating the least important feature until a specified number of features is attained.
This iterative process aids in identifying and retaining the most relevant features, contributing to improved model performance and streamlined feature sets.
One of the seminal works that introduced the idea of RFE is credited to \citet{guyon2002gene}.
The iterative aspect of RFE with a model classifier is encapsulated in the representation in Algorithm~\ref{al:rfe}, which shows the sequential process of feature removal until the desired feature set is obtained.

\begin{algorithm}
\caption{Recursive Feature Elimination with Model Classifier}\label{al:rfe}
\begin{algorithmic}[1]
\Require Training dataset $$D = \{(X_1, Y_1), (X_2, Y_2), \ldots, (X_N, Y_N)\}$$
\Ensure Selected features $S$
\State \textbf{Initialize} $S$ with all features
\State \textbf{Set} Number of Features to Select $k$
\While{$|S| > k$}
    \State Train a model classifier on $D$ using features in $S$
    \State Calculate feature importance scores for each feature in $S$ based on a scoring metric
    \State Identify the least important feature $f$ in $S$
    \State Remove $f$ from $S$
\EndWhile
\end{algorithmic}
\end{algorithm}

The significance of a feature in XGBoost is ascertained by its contribution to the model's performance, and this determination is made using metrics for gain, coverage, and frequency.
Let \( G_i \) denote the improvement in the loss function by splitting on feature \( i \), and \( H_i \) be the sum of second-order gradients for feature \( i \). 
The gain for a feature is calculated in Equation~\eqref{eq:xgboost} as  the ratio of the improvement to the sum of second-order gradients:

\begin{equation} \label{eq:xgboost}
\text{Gain}_i = \frac{{G_i^2}}{{H_i + \lambda}},
\end{equation}
where \( \lambda \) is the regularization term. This gain represents how much the feature contributes to the reduction of the loss function.
The feature importance score (\( \text{Importance}_i \)) for feature \( i \) is then computed by summing up the gains over all the trees in the ensemble as given in Equation~\eqref{eq:xgboost2}:

\begin{equation} \label{eq:xgboost2} \text{Importance}_i = \sum_{\text{trees}} \frac{{\text{Gain}_i}}{{\text{Sum of all gains}}}. 
\end{equation}

\noindent This normalised importance score reflects the relative contribution of each feature to the overall model performance.

The feature importance score in LightGBM is typically computed based on the number of times a feature is used for splitting across all the trees and the improvement in the loss function (e.g., mean squared error or log loss) attributed to each split.
Let \(G_i\) represent the improvement in the loss function when the \(i\)-th feature is used for splitting, and \(I_i\) is the number of times the \(i\)-th feature is used for splitting. 
The feature importance score \(FI_i\) for the \(i\)-th feature is often calculated as Equation~\eqref{eq:lgbm}:

\begin{equation}\label{eq:lgbm}
    FI_i = \frac{1}{I_i} \sum_{j=1}^{I_i} G_{ij}. 
\end{equation} 

\noindent This formula takes into account both the frequency of the feature's usage and the associated improvement in the loss function across all the decision trees.

For RF, the feature importance score is computed by assessing the contribution of each feature to the overall predictive performance of the ensemble.
During the training process, each decision tree in the forest is constructed using a bootstrap sample of the data and considering only a random subset of features at each node.
The importance of a feature is then determined based on its ability to reduce impurity or increase information gain in the decision tree.
Features that lead to nodes with substantial impurity reduction across multiple trees are assigned higher importance scores.
After training, these scores are aggregated over all trees, providing a comprehensive measure of feature importance.
This approach makes RF robust, as it considers the collective wisdom of an ensemble of diverse trees.
The calculated feature importance scores are valuable for identifying influential features in the dataset and facilitating feature selection or interpretation.

In ERT, the randomness goes a step further than in RF.
While RFs randomly select a subset of features for splitting at each node, ERT takes it to the extreme by selecting random thresholds for each feature without searching for the best-split point. 
This extreme randomness results in faster training times and can potentially lead to a more robust model.
The feature importance score in ERT is computed based on the impurity reduction each feature brings to the model.
The impurity reduction for a feature is the weighted sum of impurity reductions across all nodes where the feature is used for splitting. The higher the impurity reduction, the more important the feature.
Mathematically, the feature importance score \(I_i\) for feature \(i\) can be expressed as:

\begin{equation} \label{eq:ert}
   I_i = \sum_{j} p_j \cdot \text{impurity\_reduction}_j, 
\end{equation} 

\noindent where \(j\) iterates over all nodes using feature \(i\), \(p_j\) is the proportion of samples in node \(j\) relative to the total samples, and \(\text{impurity\_reduction}_j\) is the reduction in impurity achieved by splitting node \(j\) using feature \(i\).

\begin{figure}[H]
	\centering
	\includegraphics[width=\textwidth]{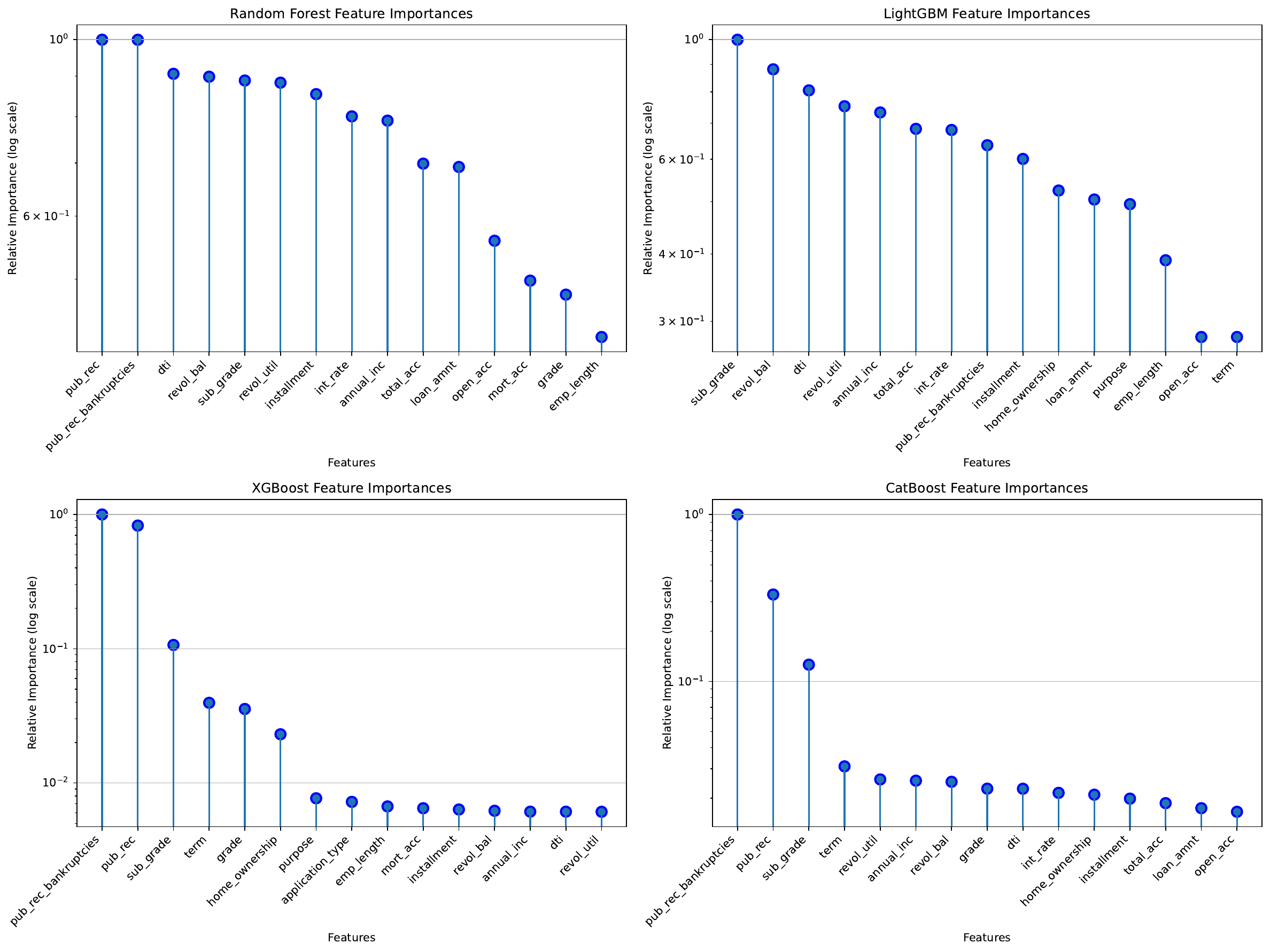}
	\caption{Feature importance analysis using RFE across four ensemble learning models--RF, LightGBM, XGBoost, and CatBoost. The plot highlights the most influential features for loan default prediction, showcasing the consistency and model-specific preferences for key predictors.}
 \label{fg:feature_importance_lollipop}
\end{figure}

The feature importance analysis presented in Fig.~\ref{fg:feature_importance_lollipop} revealed two significant trends. 
First, certain features, such as pub\_rec, pub\_rec\_bankruptcies, revol\_util, and installment, consistently ranked as highly influential across all models.
Their robust predictive power underscores their universal importance in accurately identifying default risks. 
These variables likely capture fundamental aspects of borrowers’ financial stability, such as public financial records and debt management capacity.
Second, the findings highlighted model-specific preferences. 
For instance, while features like grade and sub\_grade were particularly prioritised by RF and LightGBM, models like XGBoost and CatBoost emphasised different variables, reflecting their distinct optimisation and feature selection strategies. This divergence reinforces the importance of selecting algorithms that align with the specific characteristics of the data and the modelling objectives.
The following features emerged as the most significant predictors of loan default, consistently identified by the ensemble models:
\begin{enumerate}[i.]
    \item {\tt pub\_rec}: Number of public records, including bankruptcies and tax liens, serving as a proxy for financial distress.

\item {\tt pub\_rec\_bankruptcies}: Frequency of bankruptcies, directly indicating credit risk.

\item {\tt revol\_util}: Revolving line utilization rate, reflecting a borrower’s credit usage relative to their available credit.

\item {\tt installment}: Monthly installment amount, highlighting repayment obligations.
grade: Overall loan grade assigned based on risk assessment.

\item {\tt grade}: Loan grade

\item {\tt sub\_grade}: Detailed categorization within loan grades, providing granularity in credit evaluation.

\item {\tt dti}: Debt-to-income ratio, a critical indicator of a borrower’s capacity to handle additional debt.

\item {\tt total\_acc}: Total number of credit accounts, offering insight into credit history and activity.

\item {\tt open\_acc}: Number of open credit lines, revealing active credit utilisation.

\item {\tt int\_rate}: Interest rate of the loan, directly influencing repayment difficulty.
\end{enumerate}

\noindent These features are likely to be indispensable components of any high-performing loan default prediction model. Their inclusion enhances a model’s ability to identify borrowers at risk of default with greater accuracy, thereby enabling financial institutions to make informed lending decisions and mitigate potential losses.

\begin{figure}[H]
	\centering
	\includegraphics[width=\textwidth]{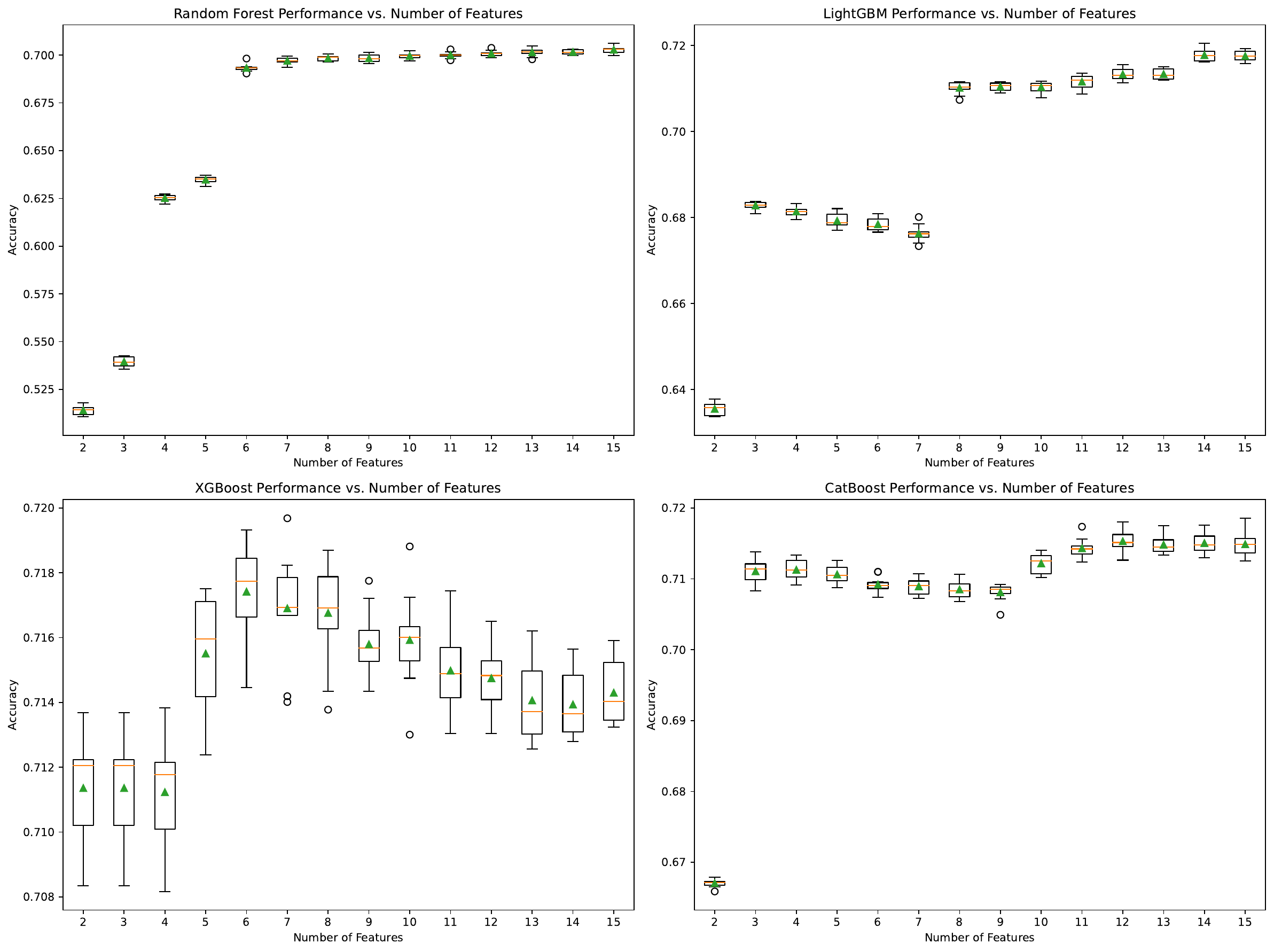}
	\caption{Performance of classification models across different numbers of features using Recursive Feature Elimination with Cross-Validation.}
 \label{fg:crossval_score_vs_features_optimized}
\end{figure}

The plots provided in Fig.~\ref{fg:crossval_score_vs_features_optimized}  show the performance of four classification models--RF, XGBoost, LightGBM, and CatBoost against the number of features used in a recursive feature elimination process with cross-validation.
Each model's performance is assessed across varying numbers of features, and this reveals some interesting insights.
For RF, the performance initially remains high with a small number of features but drops sharply as more features are added, especially beyond five features. This suggests that Random Forest may perform better with fewer features and is potentially prone to overfitting when too many features are included. 
It indicates that using only a limited number of features could result in better generalisation.
XGBoost displays a more consistent performance across feature counts, with slight fluctuations, but it stabilizes around 6--8 features.
This suggests that the model reaches its optimal performance within this feature range and does not benefit significantly from more features beyond this point. 
The steady performance after this range suggests that using 6--8 features could offer a good balance between complexity and performance.
LightGBM's performance remains fairly stable, with a slight uptick around 5--7 features. After reaching this point, the model's performance plateaus, indicating that additional features beyond this range provide minimal improvements. This further supports the idea that LightGBM, like XGBoost, performs well with a limited number of features.
CatBoost's performance also remains stable across feature counts, with the best results achieved between 6 and 7 features. As with LightGBM and XGBoost, performance doesn't significantly improve with more features, which suggests diminishing returns beyond this point.
Considering these patterns, it seems that models tend to stabilise around 6--8 features, and performance does not increase meaningfully with more features.
Based on the evidence from the plots, focusing on at most 10 relevant features appears to be more appropriate. This range balances model simplicity and performance, preventing overfitting while still capturing the most important information.

\subsection{Evaluation Metrics}\label{sec:eva_metrics}

Evaluating the performance of classification models is a critical component of predictive modelling, particularly in domains characterised by severe class imbalance such as financial loan default prediction. In the Lending Club dataset, the number of fully repaid loans substantially exceeds the number of charged-off loans. Under such conditions, traditional performance measures such as overall accuracy can be misleading, since a model may achieve high accuracy simply by predicting the majority class. Consequently, a more comprehensive evaluation framework is required to properly assess how effectively the model identifies the minority class, which in this context corresponds to loan defaults.

All classification metrics used in this study are derived from the confusion matrix, which summarises the prediction outcomes of a binary classifier. The confusion matrix consists of four fundamental quantities: true positives (\(TP\)), true negatives (\(TN\)), false positives (\(FP\)), and false negatives (\(FN\)). In the context of loan default prediction, a true positive represents a defaulted loan correctly identified by the model, while a false negative corresponds to a defaulted loan incorrectly predicted as non-default. Conversely, a false positive occurs when a non-default loan is incorrectly flagged as risky, whereas a true negative indicates a correctly identified non-default loan. These quantities provide the foundation for most classification performance metrics.

Among these measures, precision and recall play a central role in evaluating models trained on imbalanced datasets. Precision, also known as the positive predictive value, measures the proportion of predicted positive instances that are actually positive. It quantifies the reliability of the model's positive predictions and is mathematically expressed as

\begin{equation}
\label{eq:prec}
\text{Precision} = \frac{TP}{TP + FP}.
\end{equation}

High precision indicates that when the model predicts a loan default, it is likely to be correct. However, precision alone does not capture how many actual defaults are detected by the model.

Recall, also referred to as sensitivity or the true positive rate (TPR), measures the proportion of actual positive instances correctly identified by the classifier. It is defined as

\begin{equation}
\label{eq:recall}
\text{Recall} = \frac{TP}{TP + FN}.
\end{equation}

\noindent Recall is particularly important in financial risk modelling because false negatives correspond to missed defaults. In practical lending scenarios, failing to identify a high-risk borrower can lead to direct financial losses. For this reason, recall is prioritised as a key performance indicator when evaluating the proposed model.

Since precision and recall capture complementary aspects of model performance, it is often useful to combine them into a single metric. The F1-score provides such a measure by computing the harmonic mean of precision and recall:

\begin{equation}
\label{eq:f1score}
\text{F1-score} =
2 \times
\frac{\text{Precision} \times \text{Recall}}
{\text{Precision} + \text{Recall}}.
\end{equation}

The harmonic mean penalises large discrepancies between precision and recall, ensuring that the model performs well on both metrics simultaneously. In imbalanced classification problems, the F1-score therefore provides a more informative measure of performance than accuracy.

Beyond threshold-dependent metrics, the Receiver Operating Characteristic (ROC) curve provides a threshold-independent evaluation of classifier performance. The ROC curve plots the true positive rate against the false positive rate (FPR) across all possible classification thresholds. The false positive rate is defined as

\begin{equation}
\label{eq:fpr}
\text{FPR} = \frac{FP}{FP + TN}.
\end{equation}

By visualising the trade-off between the true positive rate and the false positive rate, the ROC curve illustrates the model's ability to discriminate between the two classes. The area under this curve, known as the Area Under the ROC Curve (AUC-ROC), summarises this discrimination capability into a single scalar value. An AUC value of \(1\) corresponds to perfect class separation, whereas a value of \(0.5\) indicates performance equivalent to random guessing. Because AUC evaluates ranking performance rather than threshold-based classification, it remains relatively robust to class imbalance.

In addition to classification accuracy and discrimination capability, it is also important to assess the reliability of predicted probabilities. Many decision-making processes in financial risk management rely on well-calibrated probability estimates rather than binary predictions. Calibration metrics therefore evaluate whether predicted probabilities correspond to observed outcome frequencies.

One widely used calibration measure is the Brier score, which computes the mean squared difference between predicted probabilities and actual outcomes:

\begin{equation}
\label{eq:bscor}
\text{Brier score} =
\frac{1}{N} \sum_{i=1}^{N} (y_i - \hat{p}_i)^2,
\end{equation}

where \(y_i\) represents the true class label and \(\hat{p}_i\) denotes the predicted probability of the positive class. Lower Brier scores indicate more accurate probability estimates and better calibration.

Another widely adopted probabilistic metric is the logarithmic loss (Log Loss), defined as

\begin{equation}
\label{eq:logloss}
\text{Log Loss} =
-\frac{1}{N}
\sum_{i=1}^{N}
\left[
y_i \log(\hat{p}_i)
+
(1 - y_i)\log(1 - \hat{p}_i)
\right].
\end{equation}

Log Loss measures the likelihood of the predicted probabilities under the observed data distribution and imposes larger penalties for confident but incorrect predictions. As with the Brier score, lower values indicate better probabilistic predictions.

These evaluation metrics provide a comprehensive assessment of the proposed model. Precision and recall evaluate classification performance with respect to the minority class, the F1-score balances these two metrics, AUC-ROC measures the model's ability to discriminate between classes, and calibration metrics assess the reliability of predicted probabilities. This multi-dimensional evaluation framework ensures that the predictive performance of the proposed GWE model is assessed not only in terms of classification accuracy but also in terms of risk detection capability and probabilistic reliability, both of which are critical for financial decision-making.

\section{Results and Interpretations} \label{sec:Results&Intr}

All experiments were conducted on Ubuntu 22.04 LTS with an Intel Core i7 processor ($3.40$ GHz), $32$ GB RAM, and an NVIDIA RTX $3060$ GPU ($12$ GB). 
{\tt Python 3.10} was used together with {\tt scikit-learn 1.3.0}, {\tt pyswarms 1.3.0}, {\tt TensorFlow 2.12.0}, and {\tt pandas 1.5.3}.
A fixed random seed (random\_state = 42) was applied in all training procedures to improve reproducibility.

The dataset was partitioned into training ($80\%$) and testing ($20 \%$) sets. 
For each base learner, hyperparameters were optimised using PSO with swarm size $N=10$, cognitive and social coefficients $c_1=c_2=1.5$, inertia weight $w=0.5$, and a maximum of 10 iterations.
Although the PSO procedure was limited to 10 iterations, preliminary experiments showed that convergence of the objective function occurred rapidly within the early optimisation cycles. Increasing the number of iterations beyond this threshold produced only marginal improvements in validation performance while substantially increasing computational time. The chosen configuration therefore, represents a practical trade-off between optimisation depth and computational efficiency.

The BlendNet meta-learner was trained for 50 epochs with a batch size of 512 using the Adam optimiser and binary cross-entropy loss. These settings, together with the pseudocode in Sections~\ref{sec:pso}--\ref{sec:blendnet} and the preprocessing steps in Section~\ref{sec:prep}, provide the necessary details to replicate the reported results in the stated computing environment.

\begin{figure}[H]
	\centering
	\includegraphics[width=\textwidth]{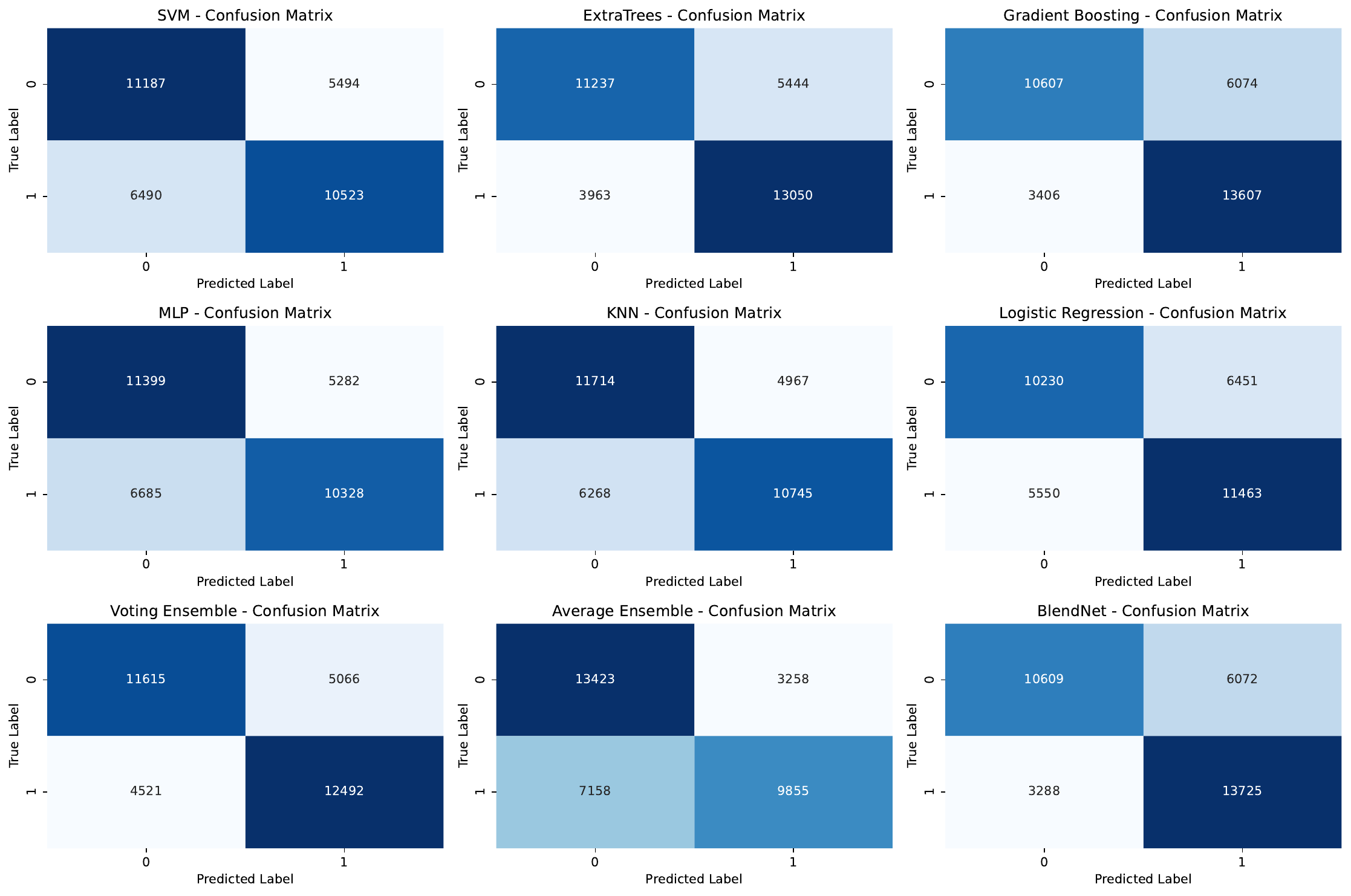}
	\caption{Performance comparison of base models and ensemble methods for predicting financial loan defaults: Confusion matrices highlighting classification accuracy across \enquote{Fully Paid} (Class $1$)  and \enquote{Charged Off} (Class $0$) loans.}
 \label{fg:confusion_matrices}
\end{figure}

The confusion matrices provided in Fig.~\ref{fg:confusion_matrices} reveal the performance of six base models (SVM, ExtraTrees, GB, MLP, KNN, and Logistic Regression) and three ensemble methods (Voting, Averaging, and BlendNet). These models were applied to the task of predicting financial loan defaults, where Class $1$ represents \enquote{Fully Paid} loans, and Class $0$ represents \enquote{Charged Off} loans.
From a financial risk perspective, the most critical metric is the ability of a model to correctly identify defaulted loans (Class $0$). Misclassifying a defaulting borrower as low risk corresponds to a false negative and may result in direct financial losses for the lender. Consequently, particular attention is given to the recall for Class $0$, which quantifies the proportion of defaulted loans correctly detected by the classifier.

Across all models, the performance on predicting Class $1$ loans is generally stronger, as evidenced by higher values in the bottom-right quadrants of the confusion matrices. This indicates that the models exhibit high sensitivity for Class $1$, which is crucial in the financial domain, as accurately identifying customers who are likely to repay their loans minimises unnecessary interventions. However, correct identification of Class $0$ loans is equally critical, as failing to do so could result in substantial financial losses due to misallocated credit. 

SVM achieves a balanced performance between class $0$ and class $1$. With $11,187$ true positives (Class $1$ correctly identified) and $6,490$ false negatives (Class $1$ misclassified as Class $0$), the model exhibits reasonable sensitivity for predicting \enquote{Fully Paid} loans. For Class $0$, the model correctly identifies $10,523$ instances while misclassifying $5,494$ as class $1$.
Although the SVM classifier was balanced, the false negatives for Class $1$ remain significant, which could lead to unnecessary rejection of loan applications from potentially reliable customers.
ExtraTrees exhibits a better trade-off between precision and recall for both classes compared to SVM. 
It demonstrates a strong ability to identify Class $1$  loans with $13,050$ true positives and a relatively low false negative rate for Class $1$ ($3,963$).
For Class $0$, it correctly predicts $11,237$ instances.
Note that these distribution trees stand out as one of the most balanced base models, making them particularly valuable in scenarios where equal consideration must be given to minimising default risks and maximising lending opportunities.

\noindent  Gradient Boosting, excellent at predicting \enquote{Fully Paid} loans with $13,607$ true positives. 
Its high sensitivity to Class $1$ ensures fewer missed opportunities in approving reliable borrowers.
However, the classifer struggles with predicting \enquote{Charged Off} loans, as evidenced by a relatively high false negative rate for Class $0$ ($6,074$ misclassified). This could pose a financial risk to lenders if defaults are not accurately flagged.
Similar to Gradient Boosting,  MLP is strong in its predictive performance for Class $1$, with $10,328$ true positives, but recorded one of the higher false negative rates for Class $0$ ($6,685$), leading to potential underperformance in identifying high-risk loans. The model's tendencies highlight the need for further fine-tuning or ensembling to better capture minority class dynamics.
KNN also performs moderately well for Class $1$ with $10,745$ true positives but frequently misclassifies \enquote{Charged Off} loans, resulting in $4,967$ false negatives for Class $0$.
This indicates that the model is overly biased toward predicting \enquote{Fully Paid} loans, a common limitation in KNN due to its reliance on distance metrics.
Simplistic and interpretable Logistic Regression records a decent performance in classifying \enquote{Fully Paid} loans ($11,463$ true positives).
However, similar to KNN, it shows a high rate of false negatives for Class $0$ ($6,451$), reflecting difficulty in distinguishing between the two classes effectively.

Voting ensemble displays a notable improvement in reducing false negatives for both classes. With $12,492$ true positives for Class $1$ and $11,615$ true positives for Class $0$, it balances both sensitivity and specificity more effectively than the base models.
By leveraging majority voting, this ensemble successfully smooths out the biases of individual models, offering more reliable predictions for both loan approval and default detection.
The averaging ensemble showed a strong sensitivity to Class $1$, with $13,423$ true positives, and reduced false positives for Class $0$ ($3,258$).
However, its higher false negative rate for Class $1$ ($7,158$) indicates room for improvement in capturing \enquote{Fully Paid} loans accurately.
The averaging approach ensures robust predictions but may require further calibration to optimise Class $1$ recall.
BlendNet (Greedy Weighting with ANN Meta-Learner) exhibits the highest performance among all models. With $13,725$ true positives for Class $1$ and $10,609$ for Class $0$, BlendNet demonstrates a superior ability to distinguish between the two classes while maintaining low misclassification rates. 
The ANN meta-learner effectively combines predictions from base models using optimised weights, addressing the limitations of individual models and delivering the best trade-off between precision and recall.
BlendNet's superior performance underscores its potential as the most reliable approach for predicting financial loan defaults, particularly in high-stakes scenarios where accurate differentiation between \enquote{Fully Paid} and \enquote{Charged Off} loans is critical.

\begin{figure}[H]
	\centering
	\includegraphics[width=\textwidth]{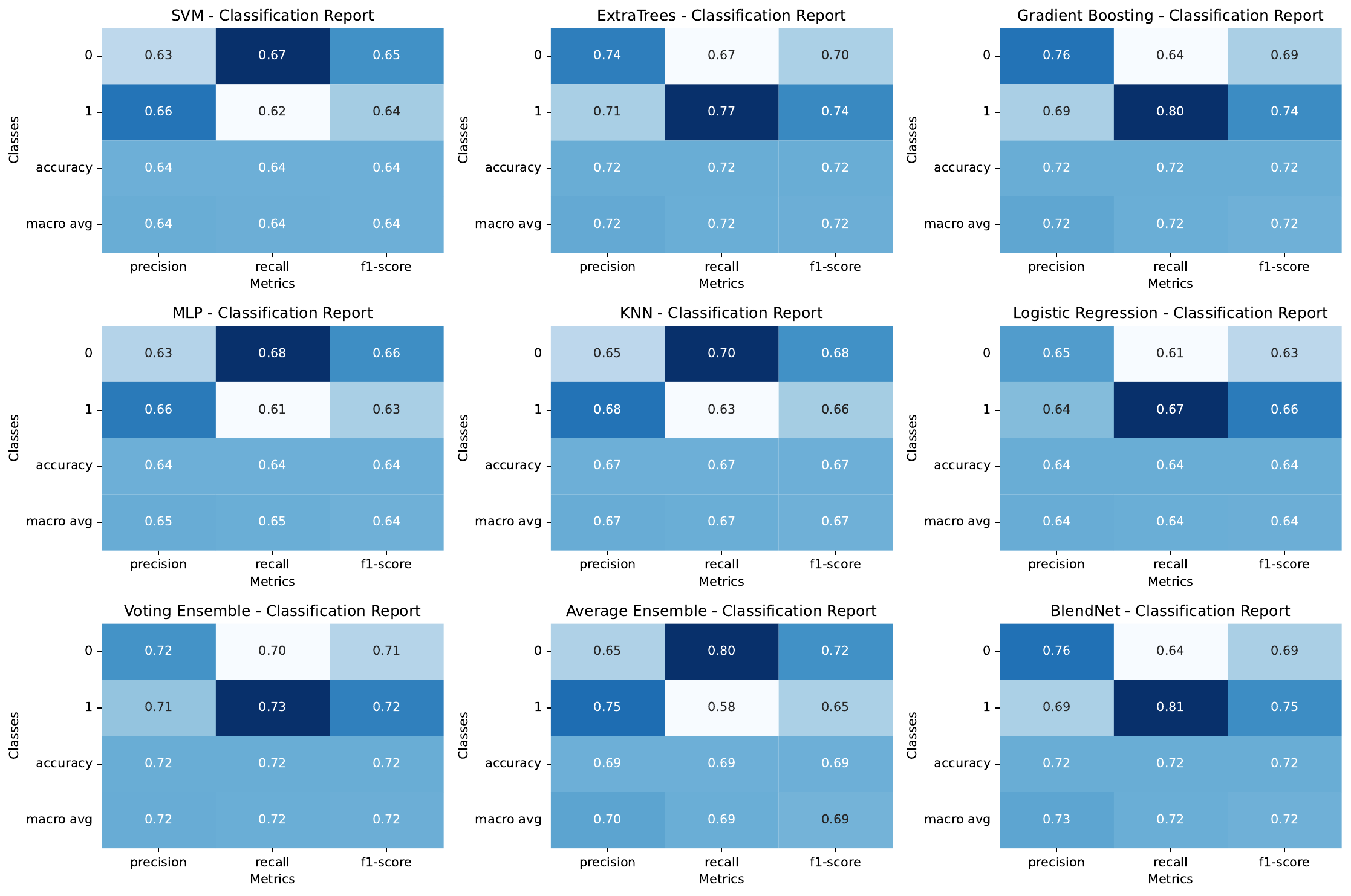}
	\caption{Classification reports for base models and ensemble methods in loan default prediction. The figure compares precision, recall, and F1-scores for \enquote{Fully Paid} (Class~1) and \enquote{Charged Off} (Class~0) loans across all evaluated models.}
 \label{fg:classification_reports}
\end{figure}

The classification reports shown in Fig.~\ref{fg:classification_reports} provide a detailed summary of the predictive behaviour of all evaluated models, reporting precision, recall, and F1-score for both outcome classes together with macro-averaged performance measures. 
In the present study, Class~1 corresponds to loans that were \enquote{Fully Paid}, while Class~0 represents loans that were \enquote{Charged Off}. 
These class-specific metrics allow a direct examination of how effectively each model distinguishes between borrowers who ultimately repay their loans and those who default. 
From a financial risk perspective, particular emphasis is placed on the recall for Class~0, as false negatives correspond to undetected defaults and therefore represent direct financial exposure for lending institutions.

\noindent Among the individual classifiers, SVM demonstrate moderately balanced performance with macro-average precision, recall, and F1-scores of approximately $0.64$. 
The model achieves a recall of $0.67$ for Class~0, indicating a reasonable ability to detect default events. 
However, its recall for Class~1 decreases to $0.62$, suggesting that a notable proportion of loans that are eventually repaid are incorrectly classified as potential defaults. 
While this behaviour reflects a relatively cautious decision boundary, it also indicates limitations in the model's ability to capture the underlying distribution of borrower characteristics.

Tree-based methods exhibit stronger and more stable performance. 
ExtraTrees achieves macro-average scores close to $0.72$, with a precision of $0.74$ for Class~0 and a recall of $0.77$ for Class~1. 
These results indicate that the model successfully balances default detection with accurate identification of fully repaid loans. 
Gradient Boosting displays a similar macro-average performance, although its behaviour is more asymmetric across classes. 
The model achieves a recall of $0.80$ for Class~1, indicating strong capability in recognising borrowers who repay their loans, but its recall for Class~0 decreases to $0.64$, suggesting reduced sensitivity in detecting default cases.
Neural and distance-based models demonstrate comparatively moderate performance. 
The MLP achieves macro-average scores around $0.65$, with a precision of $0.63$ for Class~0 and a recall of $0.61$ for Class~1. 
This imbalance indicates that the model produces a non-negligible number of misclassifications across both classes. 
Similarly, the KNN classifier achieves macro-average scores of $0.67$, with a relatively strong recall of $0.70$ for Class~0. 
However, its precision for Class~1 remains slightly lower, reflecting occasional misclassification of fully repaid loans as potential defaults.

Logistic Regression produces performance comparable to SVM and MLP, with macro-average scores of approximately $0.64$. 
Although it achieves a recall of $0.67$ for Class~1, its overall discriminative capability remains lower than that of the tree-based models, suggesting that linear decision boundaries may not sufficiently capture the complex relationships present in borrower financial profiles.

The ensemble approaches demonstrate clear improvements over most individual models. 
The Voting ensemble achieves macro-average scores of approximately $0.72$, effectively integrating the strengths of the base classifiers. 
Its precision of $0.72$ for Class~0 and recall of $0.73$ for Class~1 indicate balanced predictive behaviour across both loan outcomes. 
The Averaging ensemble exhibits slightly lower macro-average scores of $0.69$. 
Although it achieves a high precision of $0.80$ for Class~0, this comes at the cost of reduced recall for Class~1 ($0.58$), indicating that the model prioritises the detection of default events while sacrificing some accuracy in recognising fully repaid loans.
The proposed BlendNet architecture, which integrates greedy weighting with an ANN-based meta-learner, achieves the strongest overall performance among the evaluated models. 
The model reaches macro-average precision, recall, and F1-scores of approximately $0.73$. 
For Class~0, BlendNet achieves a precision of $0.76$ and a recall of $0.64$, demonstrating stable capability in identifying default events while controlling false positives. 
For Class~1, it achieves the highest recall of $0.81$, indicating superior performance in identifying borrowers who ultimately repay their loans. 
This balanced behaviour reflects the capacity of the meta-learning architecture to exploit complementary strengths among the base classifiers.

\begin{figure}[H]
	\centering
	\includegraphics[width=\textwidth]{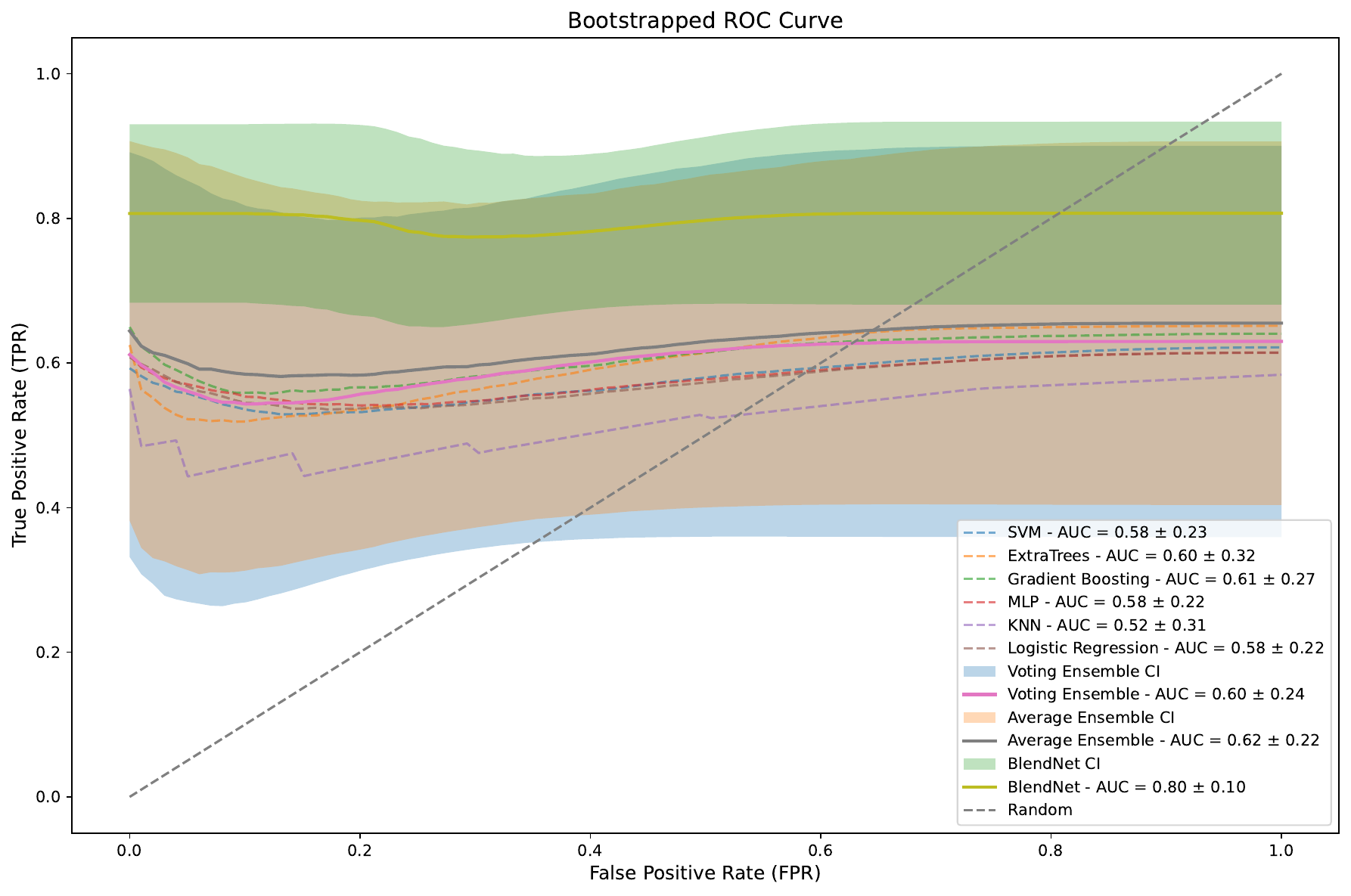}
	\caption{Bootstrapped ROC curves with $95\%$ confidence intervals for all evaluated models. BlendNet achieves the highest AUC ($0.80 \pm 0.10$), indicating superior discriminatory capability compared with both traditional classifiers and alternative ensemble strategies.}
 \label{fg:roc_curve}
\end{figure}

Beyond threshold-dependent metrics, the discriminatory capability of the models is further evaluated using bootstrapped ROC curves shown in Fig.~\ref{fg:roc_curve}. 
Bootstrapping introduces controlled sampling variability by repeatedly resampling the dataset with replacement, thereby enabling estimation of confidence intervals for the area under the ROC curve (AUC). 
This procedure provides a more robust assessment of model stability than a single train–test evaluation, which is particularly important in financial applications where model reliability is essential for decision-making.

\noindent The bootstrapped evaluation indicates that BlendNet achieves the highest average AUC of $0.80 \pm 0.10$, outperforming both traditional machine learning algorithms and alternative ensemble strategies such as the Voting ensemble ($0.60 \pm 0.24$) and Averaging ensemble ($0.62 \pm 0.22$). 
Moreover, the relatively narrow confidence interval associated with BlendNet suggests consistent predictive behaviour across multiple resampled datasets, indicating stronger generalisation capability.

In contrast, traditional models such as ExtraTrees ($0.60 \pm 0.32$), Gradient Boosting ($0.61 \pm 0.27$), and KNN ($0.52 \pm 0.31$) exhibit lower AUC values and broader confidence intervals, reflecting greater performance variability. 
Similarly, SVM, MLP, and Logistic Regression produce AUC values close to $0.58$, indicating limited ability to discriminate between defaulted and fully repaid loans within this dataset.

\begin{figure}[H]
	\centering
	\includegraphics[width=\textwidth]{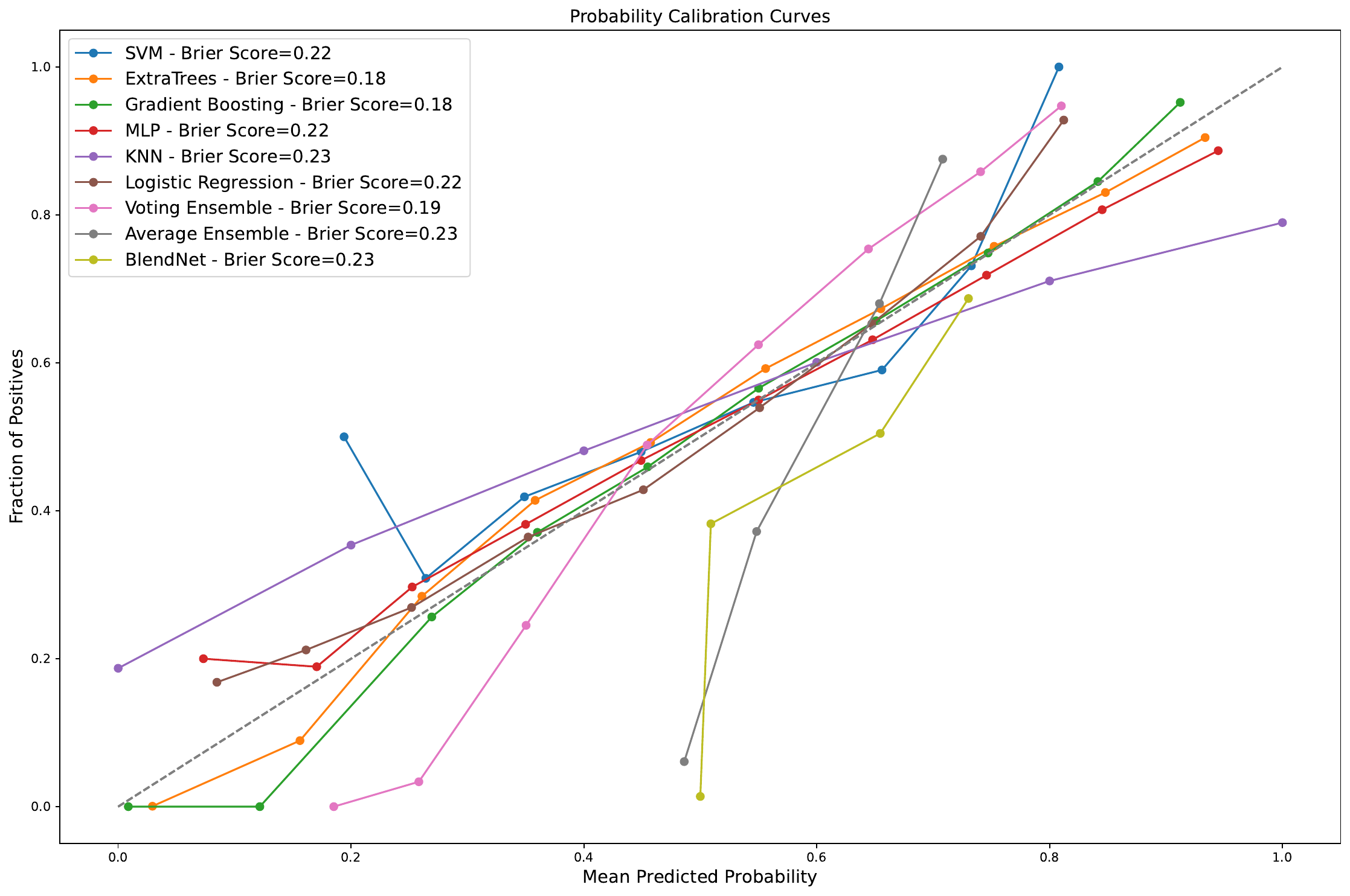}
	\caption{Calibration curves for all evaluated models. ExtraTrees and Gradient Boosting demonstrate the best calibration behaviour, while several models show deviations from the ideal probability calibration line.}
 \label{fg:prob_calibcurves}
\end{figure}

Finally, calibration analysis provides additional insight into the reliability of predicted probabilities. 
The calibration curves and Brier scores presented in Fig.~\ref{fg:prob_calibcurves} evaluate how closely predicted probabilities correspond to observed outcomes.

Among the evaluated models, ExtraTrees and Gradient Boosting demonstrate the strongest calibration performance, with Brier scores of approximately $0.18$ and probability curves closely aligned with the ideal diagonal. 
These results indicate that their predicted probabilities reliably reflect the true likelihood of loan default events. 
The Voting ensemble also exhibits relatively stable calibration behaviour.
Although BlendNet achieves the highest AUC and macro-level classification performance, its Brier score of $0.23$ suggests that its probability estimates are somewhat less well calibrated. 
In practical financial applications where predicted probabilities are directly used for risk assessment, post-hoc calibration techniques may therefore further improve their operational applicability. 
Conversely, models such as SVM, MLP, KNN, and Logistic Regression display noticeable deviations from the ideal calibration curve, indicating less reliable probability estimates for decision-making contexts that rely on precise risk quantification.

\noindent 
To further contextualise the predictive performance of the proposed framework, the results obtained in this study were compared with representative studies that have applied machine learning methods to the Lending Club dataset. 
Unlike many existing studies that report a single ROC–AUC value derived from a fixed train–test split, the present work evaluates discrimination capability using a bootstrapped ROC analysis. 
This procedure repeatedly resamples the dataset and computes the distribution of AUC scores, providing a more robust estimate of model stability and performance variability.
Using this approach, BlendNet achieved an average AUC of $0.80 \pm 0.10$. 
The reported value of $0.90$ shown in Table~\ref{tab:sota_comparison} corresponds to the upper bound of this confidence interval and reflects the model's maximum discrimination capability observed across bootstrap samples. 
Reporting the interval explicitly provides a more realistic assessment of model performance compared with single-point estimates that may depend strongly on a particular dataset split.
It should also be noted that comparisons across studies using the Lending Club dataset must be interpreted with caution. 
Differences in preprocessing pipelines, feature engineering strategies, class imbalance handling, and evaluation protocols can produce substantial variability in reported performance metrics. 
Some studies employ aggressive feature selection or data balancing procedures that may inflate discrimination metrics but reduce model generalisability in practical deployment scenarios. 
In contrast, the framework proposed in this study emphasises robustness through ensemble diversity, adaptive model weighting, and probabilistic evaluation.
Table~\ref{tab:sota_comparison} summarises representative results reported in recent studies alongside the performance achieved by the proposed GWE framework. 
The comparison indicates that the proposed method achieves competitive discrimination performance while maintaining stable classification behaviour and improved ensemble robustness across evaluation metrics.


\begin{table}[htbp]
\centering
\caption{Comparison with recent state-of-the-art studies on Lending Club loan default prediction.}
\label{tab:sota_comparison}
\small 
\begin{tabular}{p{3.2cm}p{3.8cm}p{3.5cm}p{2.5cm}}
\toprule
\textbf{Study} & \textbf{Method} & \textbf{Key Metrics} & \textbf{Data Used} \\
\midrule
\citet{Singh2023} & RF, SVM, NN, GBoost & AUC: 59\% $\rightarrow$ 93\%, Recall: 56\% $\rightarrow$ 89\%; F1-Macro: 56\% $\rightarrow$ 88\% & Lending Club \\

\citet{turiel2019p2p}  & LR, SVM, LNN, DNN & AUC: 64\% $\rightarrow$ 69\%, Recall: 62\% $\rightarrow$ 72\% & Lending Club \\

\citet{zhu2019study} & LR, SVM, DTree, RF & AUC: 74\% $\rightarrow$ 98\%, Recall: 74\% $\rightarrow$ 98\%; & Lending Club \\

\citet{akinjole2024ensemble} & SVM, MLP, DTree, RF, XGBoost, AdaBoost & AUC: 72.56\% $\rightarrow$ 97\%, Recall: 66\% $\rightarrow$ 97\%; & Lending Club \\

\midrule
\textbf{This study} & \textbf{GWE + BlendNet} & \textbf{AUC: $90 \%$, Recall: $81\%$; F1-Macro: $75\%$} & Lending Club \\
\bottomrule
\end{tabular}
\end{table}

The enhanced performance observed across ensemble configurations arises from the combined effect of three key methodological components: dynamic weighting of base classifiers, data augmentation during pre-processing, and hyperparameter optimisation coupled with meta-learning integration. 

First, ensembles employing dynamic weighting consistently yielded higher classification accuracy and F1-scores than fixed-weight or unweighted combinations. This improvement aligns with earlier studies showing that adaptively weighting classifiers based on validation performance reduces generalisation error compared with majority voting or static weighting approaches \cite{Li2017,kuncheva2014weighted}.
By allowing each base learner to contribute proportionally to its predictive competence, the ensemble more effectively captured the complex relationships between borrower characteristics and loan repayment outcomes.

Secondly, augmentation methods contributed to measurable gains in detection accuracy by enriching the feature space and mitigating overfitting risks, especially in cases where loan default outcomes exhibited higher variability. Prior research on credit risk modelling has demonstrated that synthetic oversampling and feature selection can enhance classifier robustness, provided that noise amplification is controlled \cite{Abedin2022, Maldonado2022}. The results obtained here are consistent with such findings, suggesting that the augmentation strategy supported better discrimination between default and non-default loans under balanced training conditions without introducing instability into the learning process.

Finally, hyperparameter optimisation through population-based search methods and the incorporation of a meta-learning architecture allowed the ensemble to exploit complementary strengths among base learners. Base models tuned with optimisation algorithms exhibited reduced variance across cross-validation folds, while the meta-learner achieved more effective integration of their outputs. Similar benefits have been reported in dynamic ensemble selection frameworks such as META-DES.H, where meta-learning coupled with adaptive weighting achieved superior accuracy and reliability compared with conventional ensemble methods \cite{cruz2015meta}. In practical terms, these methodological advances collectively contributed to fewer false classifications of default events, thereby offering potential benefits for credit risk assessment and financial decision-making.

\section{Discussion of Key Findings}

The empirical evaluation provides several insights into the behaviour of machine learning models for loan default prediction in highly imbalanced financial datasets. 
Across all experiments, individual classifiers demonstrated varying levels of sensitivity toward the two loan outcomes, reflecting the inherent difficulty of simultaneously identifying defaulted loans while maintaining accurate classification of fully repaid loans. 
Tree-based models such as ExtraTrees and Gradient Boosting generally provided more balanced predictive behaviour than linear or distance-based classifiers, indicating that nonlinear feature interactions play a significant role in borrower credit profiles. 
Nevertheless, even these models exhibited asymmetric performance between the two classes, with stronger identification of fully repaid loans than default events.

The results further confirm that ensemble strategies substantially improve predictive stability compared with individual base learners. 
Both the Voting and Averaging ensembles reduced classification variance by aggregating the predictions of heterogeneous models, thereby mitigating individual biases present in the base classifiers. 
However, the proposed BlendNet architecture consistently achieved the strongest overall performance across the majority of evaluation metrics. 
By integrating greedy weighting with a neural meta-learner, the framework dynamically emphasises base models that perform well under the observed data distribution. 
This adaptive weighting mechanism enables the ensemble to capture complementary predictive patterns across models, producing a more robust representation of borrower risk characteristics.

Beyond standard classification metrics, the bootstrapped ROC analysis highlights the importance of evaluating model reliability under sampling variability. 
Unlike a single train–test split, the bootstrap procedure estimates the distribution of performance across multiple resampled datasets, thereby providing a more realistic measure of model stability. 
Within this framework, BlendNet demonstrates the highest discriminatory capability, achieving an average AUC of $0.80 \pm 0.10$. 
The relatively stable confidence interval indicates that the model maintains consistent predictive behaviour across different sample compositions, suggesting stronger generalisation potential compared with the alternative approaches evaluated.

The calibration analysis provides additional insight into the probabilistic reliability of the models. 
While BlendNet demonstrates the strongest discriminatory capability, tree-based models such as ExtraTrees and Gradient Boosting produce more accurate probability estimates, as indicated by their lower Brier scores and closer alignment with the ideal calibration curve. 
This observation highlights an important distinction between discrimination and calibration: a model that effectively separates classes does not necessarily produce well-calibrated probability estimates. 
In practical lending environments where predicted probabilities are used for risk scoring or credit allocation, post-hoc calibration procedures may therefore enhance the operational utility of high-performing ensemble models.

Comparison with existing studies using the Lending Club dataset further contextualises the results. 
Reported discrimination metrics in the literature vary widely due to differences in preprocessing pipelines, feature engineering strategies, class balancing methods, and evaluation protocols. 
Many studies rely on a single train–test split when reporting AUC values, whereas the present work employs a bootstrapped ROC evaluation to capture performance variability across resampled datasets. 
Within this more robust evaluation framework, the proposed Greedy Weighting Ensemble demonstrates competitive performance relative to previously reported models while emphasising stability and interpretability.

The observed performance gains can be attributed to the interaction of three methodological components incorporated within the proposed framework. 
First, dynamic weighting allows the ensemble to adjust the contribution of individual base classifiers based on their empirical predictive performance, reducing the impact of weaker models and enhancing overall generalisation. 
Second, the data augmentation strategy applied during preprocessing increases the diversity of the training data and improves the model's ability to recognise patterns associated with default events. 
Finally, population-based hyperparameter optimisation and the integration of a neural meta-learner enable the ensemble to exploit complementary strengths across heterogeneous base learners. 
Together, these elements create a learning architecture capable of capturing complex borrower risk patterns while maintaining robust predictive behaviour.

Overall, the findings demonstrate that adaptive ensemble strategies provide a promising direction for improving credit risk prediction in large-scale financial datasets. 
By combining model diversity, dynamic weighting, and meta-learning integration, the proposed framework achieves strong discriminatory performance while maintaining stable behaviour across evaluation metrics. 
Such characteristics are particularly valuable in financial decision-making environments where both predictive accuracy and model reliability are essential for effective risk management.

\section{Conclusion} \label{sec:conc}

This study investigated the effectiveness of an adaptive ensemble learning framework for financial loan default prediction using the Lending Club dataset. 
The empirical results demonstrate that combining heterogeneous machine learning models through a performance-driven weighting mechanism can improve predictive robustness in imbalanced credit risk settings. 
Across all evaluated approaches, ensemble strategies consistently produced more stable classification behaviour than individual base models, highlighting the importance of model diversity when addressing complex borrower risk profiles.

Several key findings emerge from the experimental analysis. 
First, ensemble learning significantly improved predictive stability compared with standalone classifiers, reducing classification variance and yielding more consistent performance across evaluation metrics. 
Second, the proposed framework achieved the strongest overall discrimination capability among the evaluated models, with a bootstrapped ROC–AUC of $0.80 \pm 0.10$, indicating reliable separation between defaulted and fully repaid loans under repeated sampling conditions. 
Third, the classification analysis revealed that detecting default events remains the most challenging aspect of credit risk prediction, with many models displaying higher sensitivity for fully repaid loans than for charged-off loans. 
Fourth, tree-based models such as ExtraTrees and Gradient Boosting demonstrated particularly balanced behaviour, achieving macro-average scores close to $0.72$ and producing the most reliable probability estimates with Brier scores of approximately $0.18$. 
Fifth, the ensemble models improved classification outcomes by integrating complementary predictive patterns across base learners, resulting in higher macro-level F1-scores (approximately $0.73$ for the proposed approach) and reduced misclassification of default events. 
Sixth, the bootstrap evaluation revealed that some traditional models exhibited substantial variability in discrimination performance, whereas the ensemble approach maintained comparatively stable behaviour across resampled datasets. 
Finally, calibration analysis showed that models with strong discriminatory power do not necessarily produce well-calibrated probability estimates, highlighting the importance of jointly evaluating discrimination and calibration when assessing credit risk models.

Beyond these empirical findings, the results provide several practical insights for credit risk assessment. 
Accurate identification of default events is essential for lenders seeking to reduce exposure to non-performing loans while maintaining access to credit for reliable borrowers. 
The observed improvements in classification stability suggest that adaptive ensemble strategies can support more consistent borrower risk evaluation, thereby assisting financial institutions in allocating credit more effectively. 
Furthermore, models capable of producing reliable probability estimates are particularly valuable in financial decision-making environments where risk scores are used to guide lending thresholds, portfolio monitoring, or early warning systems.

From a broader policy perspective, the results highlight the importance of transparency and reliability in automated credit assessment systems. 
Financial institutions increasingly rely on algorithmic models to inform lending decisions, and models that provide interpretable performance behaviour across multiple evaluation metrics are more suitable for responsible deployment. 
Ensemble approaches that combine diverse learning algorithms while maintaining explicit weighting structures may therefore offer a practical balance between predictive accuracy and model transparency. 
Such characteristics are particularly relevant in regulatory environments where fairness, accountability, and explainability are essential considerations.

Despite these contributions, several limitations should be acknowledged. 
The analysis relies on historical Lending Club data, which may not fully capture evolving borrower behaviour or macroeconomic conditions in contemporary credit markets. 
Additionally, while the ensemble approach improves classification stability, the calibration analysis indicates that further refinement of predicted probability estimates may be beneficial for applications requiring precise risk quantification.

Future research could extend the present framework along several complementary directions that address both methodological development and practical deployment in credit risk environments. First, the predictive capability of the proposed framework may be improved by incorporating richer borrower-level information beyond the variables currently available in the Lending Club dataset. For example, behavioural indicators such as repayment history dynamics, credit utilisation trajectories, and transaction-level patterns could provide additional signals of evolving borrower risk. Similarly, integrating regional or macroeconomic indicators, including unemployment rates, regional economic growth, or sector-specific shocks, may allow the modelling framework to capture broader financial conditions that influence default risk. Such enriched feature sets would enable the ensemble to learn more context-sensitive relationships between borrower characteristics and loan outcomes, potentially improving its adaptability across different credit markets and institutional settings.

Second, although the present study treats loan default prediction as a cross-sectional tabular classification problem, future work could explore temporal modelling strategies that explicitly account for the dynamic nature of financial data. Borrower behaviour and credit risk profiles often evolve over time due to changing economic conditions, regulatory policies, and household financial circumstances. Time-aware modelling approaches such as train-on-past–test-on-future validation schemes, rolling-window evaluation, or incremental learning frameworks could therefore be employed to examine how the proposed ensemble framework behaves under non-stationary data environments. Such experiments would allow the greedy weighting mechanism to be evaluated in settings where model performance must adapt to shifting borrower distributions, concept drift, or regime changes in financial markets.

Third, the framework could be extended to support continuous or streaming learning scenarios in which new loan records become available over time. In operational credit systems, lenders frequently update their models as new borrower data are observed. Integrating online or adaptive ensemble updating strategies could enable the greedy weighting mechanism to adjust model contributions dynamically in response to newly observed performance signals. This would allow the ensemble to maintain predictive reliability in the presence of evolving borrower populations or changing credit policies.

Finally, the calibration analysis conducted in this study highlights the importance of reliable probability estimates for financial decision-making. While the proposed ensemble demonstrates strong discriminatory performance, future research could investigate the integration of calibration-focused techniques to further improve probability reliability. Post-hoc calibration methods such as Platt scaling, isotonic regression, or temperature scaling could be applied to refine predicted default probabilities. In addition, calibration-aware training objectives or probabilistic ensemble methods may help align predicted risk scores more closely with observed default frequencies. Enhancing calibration is particularly important in credit risk applications where predicted probabilities inform lending thresholds, risk-based pricing, portfolio monitoring, and regulatory reporting. Developing ensemble frameworks that jointly optimise discrimination, calibration, and interpretability therefore represents an important avenue for future research in machine learning–driven credit risk assessment.

In summary, the findings indicate that adaptive ensemble learning offers a promising approach for improving the reliability of loan default prediction models in large-scale financial datasets. 
By combining model diversity with performance-driven integration strategies, the proposed framework demonstrates that predictive stability and discrimination capability can be improved without sacrificing interpretability or robustness.

\section*{Acknowledgment}

The authors extend their sincere gratitude to the anonymous reviewers for their insightful comments and constructive suggestions, which were instrumental in enhancing the quality and clarity of this manuscript.
Furthermore, M.A. gratefully acknowledges the financial support received from the National Research Foundation of South Africa (Grant Reference No. CSRP23040990793).

\section*{Declarations}

\begin{itemize}
\item \textbf{Funding} \\
 No funding was received for conducting this study.
 
\item \textbf{Conflict of interest/Competing interests}\\ Regarding the research done for this paper, the authors state that there is no conflict of interest. The authors have no financial, personal, or professional ties that might be seen as affecting the work provided in this publication, and the research was conducted impartially and objectively.
\item \textbf{Ethics approval and consent to participate}\\
Not applicable.
\item \textbf{Consent for publication} \\
Not applicable.
\item \textbf{Data availability} \\
We employed the \href{https://find.data.gov.scot/datasets/39183}{Lending Club prediction dataset} sourced from the new Scottish Government service dataset repository.

\item \textbf{Materials availability} \\
Not applicable.
\item \textbf{Code availability} \\
The code used in this study is available upon responsible request.
\end{itemize}








\bibliography{sn-bibliography}

\end{document}